\documentclass{article}

\usepackage{iclr2027_conference,times}
\iclrfinalcopy

\usepackage{amsmath,amssymb,amsthm,mathtools}
\usepackage{graphicx}
\usepackage{booktabs}
\usepackage{microtype}
\usepackage{hyperref}
\usepackage{url}
\usepackage{natbib}
\usepackage{algorithm}
\usepackage{algpseudocode}
\usepackage{xcolor}
\usepackage{enumitem}
\usepackage{xspace}
\usepackage{pifont}
\usepackage{subcaption}
\usepackage{multirow}
\usepackage{colortbl}

\newcommand{\abl}[2]{#1\,{\scriptsize\textcolor{gray!80!black}{(#2)}}}

\usepackage{amsmath,amsfonts,bm}

\def\eqref#1{equation~\ref{#1}}

\def\1{\bm{1}}

\DeclareMathAlphabet{\mathsfit}{\encodingdefault}{\sfdefault}{m}{sl}
\SetMathAlphabet{\mathsfit}{bold}{\encodingdefault}{\sfdefault}{bx}{n}

\newcommand{\dH}{D_H}
\newcommand{\dFR}{d_{FR}}
\newcommand{\Lloss}{\mathcal{L}}
\newcommand{\Ldist}{\mathcal{L}_{\mathrm{distil}}}
\newcommand{\Lprox}{\mathcal{L}_{\mathrm{prox}}}

\theoremstyle{definition}

\theoremstyle{remark}

\newcommand{\method}{\textsc{GeoSD}\xspace}

\title{Geometric Self-Distillation for Reasoning Generalization}

\author{Josip Juki\'{c}\\
ILLC, University of Amsterdam\\
\texttt{j.jukic@uva.nl}
\And
Ivan Titov\\
ILLC, University of Amsterdam\\
ILCC, University of Edinburgh\\
\texttt{ititov@inf.ed.ac.uk}
}

\begin{document}

\maketitle

\begin{abstract}
On-policy distillation provides dense teacher supervision on a language model's own trajectories.
In self-distillation with privileged context, this supervision comes from the model itself, conditioned on a hint or solution trace hidden from the student.
When the teacher’s preferences hinge on privileged information, it can assign higher probability to continuations the student cannot infer from its own context.
Matching these preferences throughout training can induce predictive drift and degrade out-of-distribution (OOD) reasoning.
We propose \method{}, a self-distillation method that controls this drift through two complementary geometric terms.
A Hellinger loss weights each teacher preference by the student--teacher overlap, reducing the influence of tokens to which the student assigns low probability.
Because these influences can still accumulate, a Fisher--Rao penalty regulates predictive distance from a copy of the student refreshed periodically during training.
Both terms compare next-token distributions in Fisher--Rao geometry and are jointly optimized with a preconditioner motivated by the natural gradient.
Across three model families, \method{} retains strong in-distribution gains while improving average mathematical OOD accuracy by $5.7$--$8.6$ points over the base model.
OOD gains hold across five model scales from $1.7$B to $32$B and transfer to code generation, where \method{} improves code accuracy by $1.9$ points on average despite distilling on mathematics alone.
Our analysis of mathematical reasoning shows that standard matching rapidly concentrates probability mass at high-entropy states and that its samples confidently agree on incorrect answers. In contrast, \method{} preserves alternative token mass and reduces false consensus.
\end{abstract}


\section{Introduction}

On-policy distillation (OPD) offers a practical approach to post-training large language models (LLMs), providing dense teacher supervision on trajectories generated by the student itself \citep{agarwal2024gkd,song2026survey}.
Dense supervision is especially valuable for \emph{reasoning}, where final-answer rewards are often too sparse to reveal which intermediate steps were useful \citep{lightman2024lets}.
Rather than assigning credit only after a solution is complete, OPD uses the teacher's predictions at each state the student visits to provide token-level supervision throughout the trajectory.
In \emph{on-policy self-distillation} (OPSD), the student acts as its own teacher, conditioned on self-generated prefixes. The recipe can scale cheaply, since it requires no separate, stronger teacher.
A common and effective variant gives the teacher \emph{privileged context}, such as verified reasoning traces \citep{zhao2026selfdistilled}, textual environment feedback \citep{hubotter2026rlsd}, or other informative context available during training \citep{penaloza2026privileged,ye2026policy}. The teacher's predictions are then distilled into a student that has no access to this additional context.

Whether dense supervision helps depends on the student's ability to follow the teacher's token-level preferences.
\citet{li2026rethinking} show that OPD progresses as student and teacher assign more mass to the same high-probability tokens, while incompatible reasoning patterns can prevent effective distillation.
Privileged context makes such mismatch common: the teacher can commit to continuations that are obvious given the solution but still uncertain from the student's context \citep{harne2026privileged}.
At these mismatched states, the teacher's confident predictions can generate large gradients, even when the student assigns low probability to the same tokens.
Across many updates, the student's probability mass can concentrate on continuations it was initially uncertain about.
This shift can hurt generalization: \citet{kim2026why} observe that privileged-context OPSD suppresses uncertainty and degrades out-of-distribution (OOD) performance.

Existing approaches mostly address which teacher signals to use or how they are matched. Some filter or down-weight signals judged unreliable \citep{xing2026trust,xu2026tip,stein2026gates}; others change the direction or weighting of divergences in the Kullback--Leibler (KL) family \citep{gu2024minillm,agarwal2024gkd,ko2025distillm} or reweight token credit \citep{yang2026self}.
These choices can shape the teacher's influence at each state but do not directly regulate the \emph{predictive drift} that accumulates over training.
States that produce large updates can still provide useful supervision. Filtering them may discard that signal, while retaining them may transfer the teacher's overconfidence.

\begin{figure*}
\centering
\includegraphics[width=\textwidth]{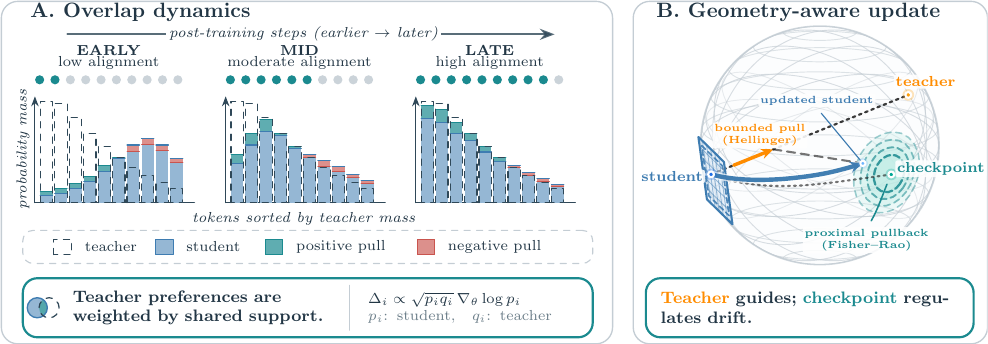}
\caption{Illustration of \method{}. \textbf{(A)} Teacher preferences are weighted by shared support. \\
\textbf{(B)} Hellinger matching guides transfer; a Fisher--Rao checkpoint penalty regulates predictive drift.}
\label{fig:overview}
\end{figure*}

We introduce \method{} (\textbf{Geo}metric \textbf{S}elf-\textbf{D}istillation),
which combines two complementary components to attenuate teacher influence at mismatched states and regulate the student's predictive drift across updates.
First, we replace the standard KL distillation objective with a \emph{Hellinger} loss.
Its gradient weights each teacher preference by the geometric mean of the student and teacher probabilities.
Teacher-preferred tokens therefore have little influence when the student assigns them little probability, but receive greater weight as the student's support increases (Figure~\ref{fig:overview}A).
Second, we penalize accumulated movement from a periodically refreshed checkpoint with the \emph{Fisher--Rao} distance between predictive distributions (Figure~\ref{fig:overview}B).
Since both terms compare next-token distributions in the same geometry, we optimize them jointly with a curvature preconditioner motivated by the natural gradient, which measures parameter updates by their effect on the student's next-token predictions.
Empirically, \method{} preserves the in-distribution (ID) gains of standard OPSD while substantially improving OOD reasoning in mathematics and code generation.
We link this improvement to how each objective reaches agreement: standard matching concentrates mass on the teacher's top choice by draining the alternatives, whereas \method{} makes that choice the most probable token while retaining more mass on alternative continuations.

Our contributions are threefold.
(i)~We introduce \method{}, a self-distillation objective that weights each teacher preference by teacher--student overlap and penalizes predictive drift from a recent checkpoint.
(ii)~\method{} improves average OOD accuracy in mathematical reasoning by $5.7$--$8.6$ points (avg@16) over the base model across three model families, whereas standard KL objectives degrade it (\S\ref{sec:experiments}).
\method{} retains ID gains and leads on all tested OOD benchmarks, with mathematical gains across five model scales (1.7B--32B) and average code gains of $1.9$ points over the base model.
(iii)~We characterize how standard matching fails out of distribution, showing that it concentrates probability mass at high-entropy states and its samples agree confidently on incorrect answers (\S\ref{sec:analysis}).

\section{Method}
\label{sec:method}

\method{} regulates the student's next-token distributions using a geometry that measures changes by their relative effect on probability mass instead of their raw magnitude.
After defining the setup (\S\ref{sec:method-setup}), we introduce overlap-aware matching to limit local teacher pressure (\S\ref{sec:method-distil}), a checkpoint penalty to control predictive drift (\S\ref{sec:method-prox}), and a curvature preconditioner to optimize both (\S\ref{sec:method-natgrad}).

\subsection{Setup}
\label{sec:method-setup}

Let $\mathcal{V}$ denote the vocabulary and $\Delta_{\mathcal{V}}$ the probability simplex over it. For an input $\mathbf{x}$ and an on-policy continuation $\mathbf{y}\sim\pi_\theta(\cdot\mid\mathbf{x})$, the student ($p$) and teacher ($q$) next-token distributions at position $t$ are
\[
    p_\theta^{(t)}=\pi_\theta(\cdot\mid\mathbf{x},\mathbf{y}_{<t}),
    \qquad
    q^{(t)}(\mathbf{c})=\pi_{\theta}(\cdot\mid\mathbf{x},\mathbf{c},\mathbf{y}_{<t})
    \ \in\ \Delta_{\mathcal{V}},
\]
where the teacher reads the same prefix but with privileged context $\mathbf{c}$, an answer-verified solution trace sampled from the base model $\pi_0$.
By default, we use the current student as the teacher with privileged context, though the objective also supports a fixed reference, a delayed copy, or an exponential moving average of the student (see Appendix~\ref{app:teacher} for evaluations).
Because $\mathbf{c}$ reveals the solution, $q^{(t)}$ provides a more informed target than $p_\theta^{(t)}$, but it can also favor tokens that the student cannot infer from $\mathbf{x}$ and $\mathbf{y}_{<t}$ alone.

We omit the parameter subscript in $q^{(t)}$ to indicate that it is a stop-gradient target. Gradients flow only through $p_\theta^{(t)}$, moving the student toward the privileged-context distribution while treating that target as fixed.
Since the default teacher shares the student's parameters, each update also changes the teacher's predictions (\S\ref{sec:method-prox}).

\begin{figure}[t]
    \centering
    \includegraphics[width=\textwidth]{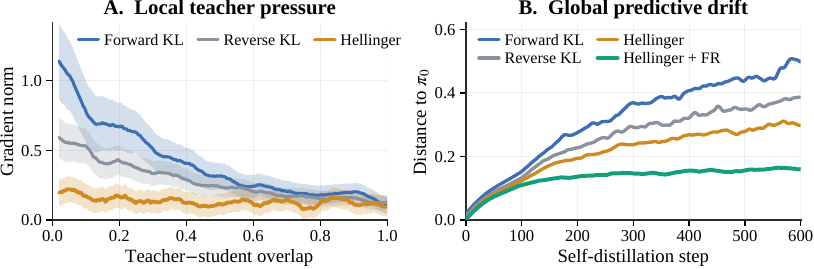}
    \caption{Local and global effects in empirical privileged-context OPSD (Qwen3-8B with full-solution teacher).
    \textbf{(A)}~Per-state gradient norm against teacher--student overlap $\sum_{i}\sqrt{p_i q_i}$.
    \textbf{(B)}~Predictive drift from the base model ($\pi_0$) over self-distillation steps, measured as $\dFR(\pi_{k}, \pi_0)$ at step $k$.}
    \label{fig:mechanism}
\end{figure}

\subsection{Local teacher influence}
\label{sec:method-distil}

The student often retains plausible alternatives beyond its top prediction, corresponding to continuations it may still sample. Suppressing one such alternative can produce only a small change in raw probability space, while effectively removing that continuation. A suitable distance should therefore measure changes relative to each token's probability mass: the same absolute change can be minor for the top prediction but significant for a lower-probability alternative.
Information geometry provides a natural way to measure these changes \citep{amari2016information,amari2000methods}. By taking the square root of each probability, we represent next-token distributions as points on a unit hypersphere, where geometric distances correspond to statistical distances between distributions (Figure~\ref{fig:overview}B; Appendix~\ref{app:sphere}).
On the sphere, two natural distances are the straight-line chord and the shortest arc along the surface.
The \emph{chord} yields the squared Hellinger divergence used for per-state matching:
\begin{equation}
\label{eq:hellinger}
\dH(p,q)=\tfrac{1}{2}\big\|\sqrt{p}-\sqrt{q}\big\|_2^2
=1-\underbrace{\sum_i\sqrt{p_i q_i}}_{\text{overlap}}.
\end{equation}
Twice the \emph{arc} length gives the Fisher--Rao distance, which we use to measure drift from a recent checkpoint (\S\ref{sec:method-prox}):
\begin{equation}
\label{eq:fr-distance}
\dFR(p,q)=2\arccos\!\Big(\sum_i\sqrt{p_i q_i}\Big).
\end{equation}

The spherical geometry does not uniquely determine the loss.
Near agreement, KL and Hellinger induce the same local geometry up to scale (Appendix~\ref{app:local-equiv}). However, their behavior is different in the low-overlap regime, which often occurs at high-entropy states in privileged-context OPSD.
We use the squared Hellinger divergence for distillation, averaged over inputs, on-policy trajectories, sampled privileged contexts $\mathbf{c}\sim\mathcal{C}(\cdot\mid\mathbf{x})$, and positions:\footnote{\label{fn:onpolicy-grad}We treat sampled trajectories as fixed and do not differentiate through $\pi_\theta(\cdot\mid\mathbf{x})$, omitting the score-function term of the outer expectation.
}
\begin{equation}
\label{eq:distillation-loss}
\Ldist(\theta)=\mathbb{E}_{\mathbf{x}\sim\mathcal{D}}\,
\mathbb{E}_{\mathbf{y}\sim\pi_\theta(\cdot\mid\mathbf{x})}\,
\mathbb{E}_{\mathbf{c}\sim\mathcal{C}(\cdot\mid\mathbf{x})}
\Big[\tfrac{1}{T}\textstyle\sum_{t=1}^{T}\dH\!\big(p_\theta^{(t)},\,q^{(t)}(\mathbf{c})\big)\Big].
\end{equation}
We choose Hellinger for its gradient. Differentiating one Hellinger term with the teacher distribution $q$ held fixed gives
\begin{equation}
\label{eq:hellinger-grad}
\nabla_\theta \dH(p_\theta,q)
=-\tfrac{1}{2}\sum_{i\in\mathcal{V}}\sqrt{p_\theta(i)\,q_i}\;\nabla_\theta\log p_\theta(i),
\end{equation}
which scales each token's influence by the geometric-mean overlap
$\sqrt{p_\theta(i)\,q_i}$. The weight vanishes as $p_\theta(i)\!\to\!0$, so even a token the teacher is certain of receives little weight when the student assigns it negligible probability.
The influence of each teacher preference depends on the student's existing support as well as the teacher's confidence.

Figure~\ref{fig:mechanism}A shows that forward-KL gradient norms grow as overlap falls: weighting by teacher mass $q_i$ maintains pressure even on tokens the student barely supports.
Such mismatches are common in student rollouts (Figure~\ref{fig:overlap_density}, Appendix~\ref{app:teacher-top-token-support}).
Reverse KL weights its log-ratio signal by student mass $p_\theta(i)$, which attenuates unsupported signals but makes it mode-seeking.
Hellinger's geometric-mean weighting attenuates low-support signals while retaining influence from overlapping alternatives.

\subsection{Global drift control}
\label{sec:method-prox}
Overlap weighting limits teacher influence at each state, but small updates can still accumulate into substantial predictive drift over training.
Because the student and teacher share parameters, self-distillation can compound this effect: shifts in the student's predictions carry over to the teacher's later targets.

Penalizing displacement from a reference model is an established practice in post-training (\S\ref{sec:rw}). We adopt this approach using the squared Fisher--Rao distance from a recent checkpoint $\theta_{\mathrm{ckpt}}$, refreshed every $K_{\mathrm{ckpt}}$ steps:\footref{fn:onpolicy-grad}
\begin{equation}
\label{eq:prox-term}
\Lprox(\theta;\theta_{\mathrm{ckpt}})=\mathbb{E}_{\mathbf{x},\,\mathbf{y}\sim\pi_\theta}
\Big[\tfrac{1}{T}\textstyle\sum_{t=1}^{T}
\dFR^2\!\big(\pi_\theta(\cdot\mid\mathbf{x},\mathbf{y}_{<t}),\,
\pi_{\theta_{\mathrm{ckpt}}}(\cdot\mid\mathbf{x},\mathbf{y}_{<t})\big)\Big].
\end{equation}
Refreshing the checkpoint $\theta_{\mathrm{ckpt}}$ regularly allows continued adaptation that a fixed anchor at $\pi_0$ could restrict.
In our experiments, overlap weighting alone slows cumulative drift but does not stop it, and adding the proximal term keeps it controlled (Figure~\ref{fig:mechanism}B).

\subsection{Full objective and preconditioned update}
\label{sec:method-natgrad}
The objective combines the two terms,
\begin{equation}
\label{eq:full-loss}
\Lloss(\theta)=\Ldist(\theta)+\lambda\,\Lprox(\theta;\theta_{\mathrm{ckpt}}),
\end{equation}
with $\lambda$ setting the strength of the drift penalty.
Both terms measure differences between next-token distributions in the same geometry. We use this geometry to guide the parameter update as well.
A \emph{natural gradient} gives the direction of greatest local decrease in the objective for a fixed, small change in the student's predictive distribution, measured by the Fisher metric.
This amounts to preconditioning the gradient with the inverse Fisher information matrix $F_k^{-1}$ of the current policy \citep{amari1998natural,schulman2015trust}:
\begin{equation}
\label{eq:nat-grad}
\theta_{k+1}=\theta_k-\eta\,F_k^{-1}\,\nabla_\theta\Lloss(\theta_k).
\end{equation}
Under a local approximation, the proximal term contributes a pull toward the checkpoint, weighted by the checkpoint's Fisher information and preconditioned by the current policy's inverse Fisher (Appendix~\ref{app:pullback}).

At LLM scale the full Fisher is infeasible to form or invert, so we instead precondition with a damped Kronecker-factored curvature estimate (K-FAC; \citealp{martens2015kfac}). This layerwise approximation avoids materializing the full matrix and reuses the existing backward pass (Appendix~\ref{app:exp-kfac}).
It keeps the update curvature-aware while adding a modest cost in memory ($21$\% total, $8$\% on training workers at 8B) and runtime ($1.1\times$ wall-clock) over a standard distillation step (Appendix~\ref{app:compute}).
\section{Experiments}
\label{sec:experiments}

\subsection{Experimental setup}
\label{sec:setup}

\paragraph{Models.}
We evaluate along two independent axes. For \emph{scale}, we use the Qwen3 dense family at five sizes: 1.7B, 4B, 8B, 14B, and 32B \citep{yang2025qwen3}.
For \emph{cross-family generalization}, we compare three models from different families: Qwen3-8B \citep{yang2025qwen3}, Olmo-3-7B-Think \citep{olmo2025olmo}, and DeepSeek-R1-Distill-Llama-8B (DS-R1-Llama-8B; \citealp{guo2025deepseek}). At 14B and 32B, we use a more memory-efficient K-FAC configuration (Appendix~\ref{app:exp-kfac}).

\paragraph{Training and evaluation.}
On-policy trajectories are sampled from the student on problems from DAPO-Math-17k~\citep{yu2026dapo}, a competition-mathematics dataset with verifiable integer answers but no reference solutions. In-distribution performance is measured on a held-out split.
We first evaluate generalization on related mathematics benchmarks, then test transfer to code generation.
For mathematics, we use AIME~2024, AIME~2025, AMC~2023, and MATH-500~\citep{lightman2024lets}, where prior work reports degradation from privileged-context OPSD~\citep{kim2026why}. These benchmarks all cover competition mathematics, so they mainly test shifts in problem source and difficulty rather than domain.
We report avg@$k$ (average accuracy over $k$ samples) and pass@$k$ (the fraction of problems with at least one correct answer among $k$ samples).
By default, the privileged context is a full answer-verified solution, following the OPSD convention~\citep{zhao2026selfdistilled, kim2026why}. We construct solutions offline by sampling up to $32$ rollouts per problem from $\pi_0$ and retaining those with verified answers. Problems with no verified solution are excluded from training. At each training step, we draw one retained solution as the teacher's privileged context.
To assess cross-domain transfer, we evaluate Qwen3-8B trained only on DAPO-Math-17k on HumanEval+~\citep{chen2021humaneval, liu2023your}, MBPP+~\citep{austin2021mbpp, liu2023your}, and LiveCodeBench~\citep{jain2025lcb}.
For both domains, aggregate OOD accuracy is the unweighted mean across the corresponding benchmarks.
For the main comparisons, we report \text{avg}@16 averaged over ten training and evaluation runs with different random seeds.
Full data, rollout, and evaluation settings are in Appendix~\ref{app:exp-details}.

\paragraph{Baselines.}
We compare \method{} against three baseline families.
\emph{Reference baselines} include \textbf{Base} ($\pi_0$; initial model checkpoint), \textbf{SFT} (supervised fine-tuning) on privileged teacher traces, and \textbf{GRPO}~\citep{shao2024deepseekmath} with verifiable rewards on the same training prompts. These respectively represent the pre-trained reference point, off-policy imitation of privileged solutions, and a privilege-free sparse-reward post-training baseline.
\emph{Divergence controls} use the same data and rollout pipeline with alternative per-position losses: forward KL (\textbf{FwdKL}; \citealp{hinton2015kd, zhao2026selfdistilled}), reverse KL (\textbf{RevKL}; \citealp{gu2024minillm}), Jensen--Shannon (\textbf{JSD}, $\beta{=}0.5$; \citealp{agarwal2024gkd}), and skewed KL (\textbf{SkewKL}, $\alpha{=}0.1$; \citealp{ko2024distillm,ko2025distillm}).
\emph{Filtering baselines} control which teacher signals to use: \textbf{TrOPD}~\citep{xing2026trust} separates teacher-reliable regions from outlier regions using teacher--student agreement, while \textbf{TIP}~\citep{xu2026tip} selects informative tokens using student uncertainty and teacher--student mismatch.
We provide implementation details for baselines in Appendix~\ref{app:exp-baselines}.

\paragraph{Compute overhead.}
\method{} requires only $1.10\times$ the wall-clock time of standard OPSD, as checkpoint scoring overlaps with rollout generation.
The checkpoint forward pass and K-FAC accumulation raise optimizer FLOPs to $1.26\times$, and the checkpoint weights and K-FAC state raise total memory to $1.21\times$ the baseline.
Since the checkpoint resides on inference workers by default, training-worker memory grows by only ${\approx}8\%$. A detailed breakdown is in Appendix~\ref{app:compute}.

\subsection{Generalization under privileged self-distillation}
\label{sec:exp-generalization}

\begin{table}[t]
\centering
\small
\caption{ID and OOD accuracy (avg@16) across three model families.
OOD math is the unweighted average over AIME 2024, AIME 2025,
AMC 2023, and MATH-500; OOD code is the unweighted average over
HumanEval+, MBPP+, and LiveCodeBench, evaluated zero-shot
without code supervision during post-training.
Values are means over ten runs, with sample standard deviations in subscripts;
aggregate scores are computed within each run before calculating
their means and standard deviations.
$\Delta_{\mathrm{math}}$ is the change in OOD math relative to $\pi_0$.
Best in bold; second-best underlined among reported methods.}
\label{tab:main_id_ood}
\setlength{\tabcolsep}{3.5pt}
\resizebox{\textwidth}{!}{%
\begin{tabular}{l ccrc ccr ccr}
\toprule
& \multicolumn{4}{c}{Qwen3-8B}
& \multicolumn{3}{c}{Olmo-3-7B-Think}
& \multicolumn{3}{c}{DS-R1-Llama-8B} \\
\cmidrule(lr){2-5}
\cmidrule(lr){6-8}
\cmidrule(lr){9-11}
Method
& ID & OOD math & $\Delta_{\mathrm{math}}$ & OOD code
& ID & OOD math & $\Delta_{\mathrm{math}}$
& ID & OOD math & $\Delta_{\mathrm{math}}$ \\
\midrule
\multicolumn{11}{l}{\emph{Reference baselines}} \\
$\pi_0$
& $61.4_{0.4}$ & $51.9_{0.7}$ & -- & $46.6_{0.4}$
& $53.3_{0.7}$ & $47.2_{0.9}$ & --
& $49.6_{0.5}$ & $43.7_{1.2}$ & -- \\
SFT
& $80.8_{2.5}$ & $39.4_{1.2}$ & $-12.5$ & $41.8_{0.7}$
& $73.8_{3.0}$ & $42.8_{1.9}$ & $-4.4$
& $61.8_{3.3}$ & $38.9_{2.2}$ & $-4.8$ \\
GRPO
& $80.9_{1.6}$ & $53.6_{1.2}$ & $+1.7$
& $\underline{47.3}_{0.4}$
& $74.7_{2.0}$ & $47.6_{1.0}$ & $+0.4$
& $62.5_{1.8}$ & $46.2_{1.2}$ & $+2.5$ \\
\midrule
\multicolumn{11}{l}{\emph{Divergence controls}} \\
FwdKL
& $\mathbf{84.1}_{2.4}$ & $43.8_{1.1}$ & $-8.1$
& $42.7_{0.7}$
& $75.1_{2.8}$ & $42.6_{1.5}$ & $-4.6$
& $\mathbf{70.1}_{3.1}$ & $37.8_{1.7}$ & $-5.9$ \\
RevKL
& $81.8_{2.1}$ & $44.4_{1.3}$ & $-7.5$
& $44.3_{0.9}$
& $\mathbf{77.2}_{2.4}$ & $43.9_{1.6}$ & $-3.3$
& $68.3_{1.6}$ & $40.8_{1.2}$ & $-2.9$ \\
JSD
& $80.7_{1.9}$ & $51.8_{0.9}$ & $-0.1$ & $46.5_{0.5}$ 
& $73.1_{2.1}$ & $46.1_{0.7}$ & $-1.1$
& $66.7_{1.5}$ & $42.9_{1.1}$ & $-0.8$ \\
SkewKL
& $82.0_{1.8}$ & $52.7_{1.2}$ & $+0.8$ &  $47.0_{0.6}$
& $70.6_{1.7}$ & $47.9_{0.8}$ & $+0.7$
& $65.1_{2.3}$ & $41.5_{1.5}$ & $-2.2$ \\
\midrule
\multicolumn{11}{l}{\emph{Filtering baselines}} \\
TrOPD
& $81.6_{1.5}$ & $\underline{54.8}_{1.1}$
& $\underline{+2.9}$ & $47.2_{0.8}$
& $74.9_{1.7}$ & $\underline{50.4}_{0.7}$
& $\underline{+3.2}$
& $65.2_{2.0}$ & $44.1_{0.7}$ & $+0.4$ \\
TIP
& $76.9_{1.4}$ & $53.3_{0.9}$ & $+1.4$ & $46.8_{0.3}$
& $73.6_{1.8}$ & $50.1_{1.0}$ & $+2.9$
& $66.0_{1.9}$ & $\underline{46.8}_{1.0}$
& $\underline{+3.1}$ \\
\midrule
\method{}
& $\underline{82.7}_{0.7}$ & $\mathbf{60.5}_{0.5}$
& $\mathbf{+8.6}$ & $\mathbf{48.5}_{0.2}$
& $\underline{75.5}_{1.1}$ & $\mathbf{54.8}_{0.6}$
& $\mathbf{+7.6}$
& $\underline{69.4}_{1.3}$ & $\mathbf{49.4}_{0.6}$
& $\mathbf{+5.7}$ \\
\bottomrule
\end{tabular}%
}
\end{table}

\paragraph{Geometry improves the ID--OOD reasoning frontier.}
Table~\ref{tab:main_id_ood} compares post-training methods on ID and OOD performance, showing that large in-distribution gains do not necessarily transfer under distribution shift.
FwdKL achieves some of the largest ID gains of any method, yet its OOD accuracy degrades relative to $\pi_0$ in all three families by $4.6$--$8.1$ points avg@$16$.
This combination of ID improvement and OOD degradation is consistent with prior reports on imitation-style post-training.
Softer objectives (SkewKL, JSD) reduce the degradation but yield little or no improvement over the base model, and the filtering baselines (TrOPD, TIP) yield only modest OOD gains ($+0.4$ to $+3.2$).
In contrast, \method{} improves average OOD accuracy by $5.7$--$8.6$ points avg@$16$ over $\pi_0$ while retaining strong ID gains; it ranks first on OOD in every family while staying within $0.7$--$1.7$ points of the best ID score.
Moreover, \method{} achieves the highest OOD accuracy on each of the four benchmarks across all three model families (see Appendix~Table~\ref{tab:ood_per_bench} for the per-benchmark breakdown).
Its improvement in average OOD accuracy over the strongest baseline is statistically significant in every evaluated model family (Appendix~Table~\ref{tab:sig}).
To make this trade-off easier to compare visually, we plot ID and OOD gains together in Appendix~Figure~\ref{fig:frontier}, with marker size indicating OOD pass@16. In every family, baselines that match or exceed \method{} on ID accuracy fall well below it on OOD accuracy.
This separation also persists across learning-rate and early-stopping sweeps, where weakening or shortening FwdKL training recovers OOD accuracy at the expense of ID gains, without exceeding the base model's OOD performance.
\method{}, by comparison, improves both ID and OOD accuracy across the evaluated configurations (Appendix~\ref{app:stepsize}).
These comparisons show that smaller updates or earlier stopping alone do not match the combination of ID performance and OOD gains achieved by \method{}.

\paragraph{Transfer from mathematics to code generation.}
We also evaluate code generation with Qwen3-8B after distilling exclusively on DAPO-Math, without any code supervision.
As shown in Table~\ref{tab:main_id_ood}, \method{} improves average OOD code accuracy from $46.6$ to $48.5$, a gain of $1.9$ percentage points over $\pi_0$ and $1.2$ points over GRPO, the strongest evaluated code baseline.
SFT, FwdKL, and RevKL instead reduce average code accuracy, while JSD makes little difference, and SkewKL, GRPO, and the filtering baselines yield only small gains.
The improvement over $\pi_0$ holds on all three benchmarks: HumanEval+, MBPP+, and LiveCodeBench (App.~Table~\ref{tab:code}).
Although smaller than the gains in mathematics, the consistent improvements in code generation suggest that regulating predictive drift during mathematics distillation may preserve reasoning capabilities useful in other domains.

\begin{figure}[t]
    \centering
    \begin{minipage}[t]{0.49\linewidth}
        \vspace{0pt}
        \centering
        \includegraphics[width=\linewidth]{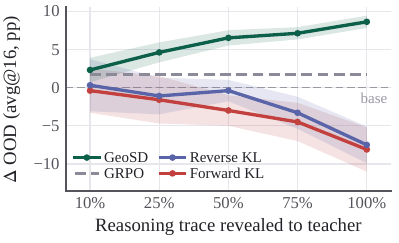}
        \caption{\textbf{Privilege sweep.} Qwen3-8B OOD
        accuracy change (avg@$16$; vs.\ $\pi_0$) as a
        function of the solution fraction revealed to the teacher.}
        \label{fig:privilege}
    \end{minipage}\hfill%
    \begin{minipage}[t]{0.49\linewidth}
        \vspace{0pt}
        \centering
        \includegraphics[width=\linewidth]{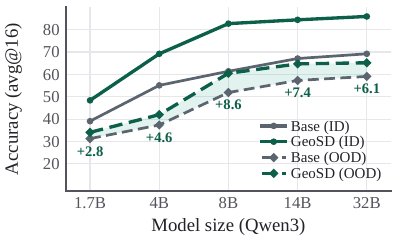}
        \caption{\textbf{Scale.} ID and OOD accuracy
        (avg@$16$) across Qwen3 model sizes
        (1.7B--32B) for the base model and \method{}.}
        \label{fig:scale}
    \end{minipage}
\end{figure}

\paragraph{From partial to full solution context.}
To test how the amount of privileged information affects generalization, we split each reference solution into sentences and reveal progressively longer prefixes to the teacher, up to the full solution, holding other settings fixed.
Figure~\ref{fig:privilege} shows that FwdKL degrades even when only a small fraction of the solution is revealed, while RevKL remains near the base model until roughly half the solution is revealed, then declines sharply. Both FwdKL and RevKL reach their worst OOD accuracy at full privilege.
In contrast, \method{} improves monotonically across the same range.
This pattern suggests that richer supervision is more useful when teacher preferences are weighted by the student's current predictions, limiting pressure toward continuations to which the student assigns low probability.

\paragraph{Geometric control complements scale.}
Figure~\ref{fig:scale} examines \method{} across the Qwen3 ladder, from 1.7B to 32B. Larger base models are stronger on both ID and OOD, and \method{} further improves OOD accuracy over the base model at every size.
The gain is non-monotone in scale, peaking at $+8.6$ points at 8B and remaining substantial at $+6.1$ at 32B, still larger than the gains at 1.7B and 4B.
Even for larger models, geometric regulation brings clear improvements during self-distillation.

\begin{table}[t]
\centering
\small
\caption{Ablations of the loss, proximal penalty, and optimizer on Qwen3-8B. Cells report OOD avg@16 (ID avg@16), each a mean over three training runs with different seeds (configurations from Table~\ref{tab:main_id_ood} are carried over). Penalties: none, Euclidean parameter distance to the checkpoint ($L_2(\theta)$), or Fisher--Rao (FR).
Remaining settings follow Table~\ref{tab:hparams}. Bottom-right cell in gray is full \method{}.}
\label{tab:ablation}
\begin{tabular}{l ccc ccc}
\toprule
& \multicolumn{3}{c}{AdamW} & \multicolumn{3}{c}{Preconditioner (K-FAC)} \\
\cmidrule(lr){2-4}\cmidrule(lr){5-7}
Loss & none & $L_2(\theta)$ & FR & none & $L_2(\theta)$ & FR \\
\midrule
Forward KL & \abl{43.8}{84.1}   & \abl{47.2}{83.6} & \abl{48.9}{83.1} & \abl{45.1}{84.4} & \abl{48.3}{83.9} & \abl{50.6}{83.3} \\
JSD        & \abl{51.8}{80.7}   & \abl{52.9}{80.5} & \abl{53.8}{80.3} & \abl{52.5}{81.0} & \abl{54.0}{80.6} & \abl{56.1}{80.2} \\
Hellinger  & \abl{52.4}{82.7} & \abl{54.8}{82.4} & \abl{56.7}{82.1}   & \abl{53.8}{83.3}   & \abl{57.2}{83.0} & \cellcolor{gray!12}\abl{\textbf{60.5}}{82.7} \\
\bottomrule
\end{tabular}
\end{table}

\subsection{Ablations}
\label{sec:exp-ablations}
\method{} couples three components: overlap-aware matching, a proximal penalty on cumulative drift, and a K-FAC preconditioned update.
Table~\ref{tab:ablation} separates their contributions on Qwen3-8B. Replacing Hellinger with JSD, removing the proximal penalty, or replacing K-FAC with AdamW lowers OOD accuracy from $60.5$ to $56.1$, $53.8$, and $56.7$, respectively. Across the full grid, Fisher--Rao regularization outperforms a Euclidean parameter penalty, and K-FAC improves OOD accuracy in every matched loss--penalty comparison. Yet forward KL with both Fisher--Rao regularization and K-FAC remains below the base model ($50.6$ versus $51.9$), indicating that regularization and preconditioning alone do not recover the full gain.

Simpler ways to limit teacher influence do not recover the same trade-off (Appendix~\ref{app:controls}). Softening the forward-KL teacher improves OOD retention but sacrifices ID accuracy and leaves OOD accuracy below $\pi_0$ at every tested temperature. Even with Fisher--Rao regularization, forward KL with per-state gradient clipping and Hellinger with clipped AdamW both fall well short of the full method on OOD accuracy. Fixing the anchor at $\pi_0$ further reduces drift but lowers both ID ($82.7 \to 71.8$) and OOD ($60.5 \to 55.6$) accuracy.
OOD gains over the base model persist across teacher choices and tested hyperparameters, with little variation among moving teachers under the full objective and a benefit from proximal regularization even when the teacher is frozen (Appendices~\ref{app:teacher} and~\ref{app:sensitivity}).

\section{Alignment without collapse}
\label{sec:analysis}

Why does standard OPSD improve ID performance while hurting generalization?
To explore this, we analyze concentration in next-token distributions at high-entropy states (\S\ref{sec:high-ent}) and agreement among final answers from different rollouts (\S\ref{sec:false-consensus}).
We find that rapid concentration at individual token positions is accompanied by strong agreement on incorrect final answers.

\subsection{Local concentration at high-entropy states}
\label{sec:high-ent}

We examine states where the base model is least certain and privileged context may substantially change its predictions.
For each rollout we select the top $10\%$ of positions by the base model's ($\pi_0$) next-token entropy.
We hold these positions fixed across methods and checkpoints, ensuring that comparisons are not affected by changes in which states each model treats as uncertain.
At each selected position we rank the candidate tokens by the teacher's probability, with rank~1 the teacher's most probable token, and track how the student's average mass on ranks $1$--$8$ evolves over training.

\begin{figure}[t]
    \centering
    \includegraphics[width=\textwidth]{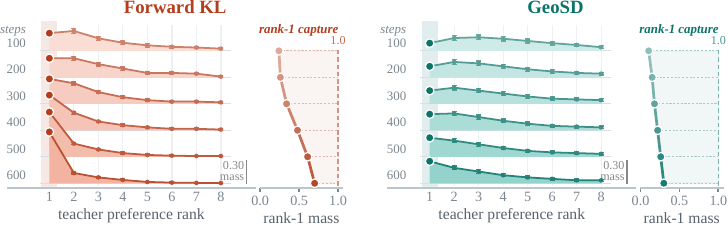}
    \caption{%
        \textbf{The shape of teacher--student alignment.} Student next-token mass at high-entropy decision points (top $10\%$ by $\pi_0$ entropy) for Qwen3-8B. At each position, tokens are sorted by the teacher's preference rank; each ridge shows the student's average mass over these ranks.
    }
    \label{fig:alignment}
\end{figure}

Figure~\ref{fig:alignment} compares FwdKL and \method{} across post-training checkpoints.
Under forward KL, rank-1 mass increases sharply over training, concentrating probability on the teacher's top choice.
Under \method{}, rank-1 mass rises gradually and substantial mass remains on lower-ranked teacher tokens even after the top choice becomes the student's modal token.
Both methods therefore increase agreement with the teacher's top choice, but \method{} retains more probability mass on alternatives.
The rapid concentration on the top choice, i.e., \emph{rank-1 collapse}, leaves less probability mass for alternatives. When lower-ranked teacher tokens represent viable continuations, this makes paths that may be useful out of distribution less likely to be explored.
These token-level differences leave open whether rollouts converge on correct answers or shared mistakes. We next examine agreement and correctness at the answer level.

\subsection{How concentration propagates: false consensus}
\label{sec:false-consensus}

\begin{figure}[t]
    \centering
    \includegraphics[width=\textwidth]{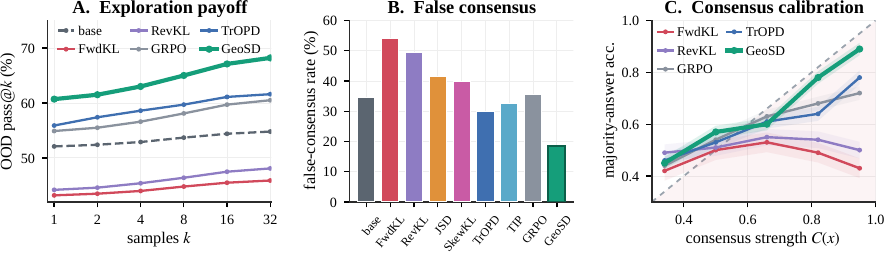}
    \caption{
        \textbf{Sampling and false consensus.}
        \textbf{(A)}~OOD pass@$k$ versus number of samples $k$ in a single run.
        \textbf{(B)}~False-consensus rate: how often samples agree strongly $P(C(x)\geq \tau \mid \hat a(x)\neq a^\star)$, conditioned on a wrong majority answer.
        \textbf{(C)}~Majority-answer accuracy versus consensus strength $C(x)$: whether stronger agreement signals a more reliable answer.
    }
    \label{fig:consensus}
\end{figure}

For each OOD problem $x$, we draw $m=16$ solutions and record their final answers.
For any answer $a$, $\hat{p}_x(a)$ denotes the fraction of samples that produce it. We denote the most frequent answer by $\hat{a}(x)$ and define the \emph{consensus strength} $C(x)$ as the fraction of samples agreeing on that answer:
\begin{equation}
\hat{p}_x(a) = \frac{1}{m}\sum_{j=1}^{m}\mathbf{1}[a_j=a],
\qquad
C(x) = \max_a \hat{p}_x(a),
\qquad
\hat{a}(x) = \arg\max_a \hat{p}_x(a).
\end{equation}
A problem $x$ exhibits \emph{false consensus} when its sampled solutions agree strongly on an incorrect answer, $C(x)\geq\tau$ and $\hat{a}(x)\neq a^\star(x)$, where $a^\star(x)$ is the correct answer and $\tau=0.75$.
This distinguishes productive self-consistency, where samples agree on the correct answer, from false consensus, where they agree strongly on an incorrect one.

Figure~\ref{fig:consensus} compares the benefits of additional sampling, the frequency of false consensus, and the relationship between agreement and correctness on OOD problems. 
FwdKL remains below the base model at every evaluated sample budget, gaining just $3.5$ points in pass@$k$ from $k=1$ to $k=32$, compared with $7.5$ points for \method{}, which also starts from higher accuracy (Panel~A).
Repeated samples can also agree strongly on an incorrect answer. Conditioning on an incorrect majority answer to reduce the accuracy confound, false consensus is most frequent under FwdKL and least frequent under \method{} (Panel~B).
How well agreement predicts correctness also depends on the method. As consensus increases through the bins above $C(x)=0.6$, majority-answer accuracy rises from $0.60$ to $0.89$ for \method{}, but falls from $0.53$ to $0.43$ for FwdKL (Panel~C).
Keeping plausible alternatives available during distillation may therefore help the student find correct answers while making agreement a more reliable sign of correctness.
\section{Related work}
\label{sec:rw}

Privileged-context self-distillation uses richer training-time contexts to supervise the student on its own rollouts \citep{zhao2026selfdistilled,ye2026policy,penaloza2026privileged}.
This asymmetry can suppress uncertainty and degrade OOD reasoning \citep{kim2026why}.
Similarly, \citet{harne2026privileged} show that privileged teachers can favor reference-specific trajectories and penalize exploratory tokens, reducing distillation loss without improving task accuracy.

Existing approaches address two aspects of teacher supervision. Signal selection uses reliability~\citep{xing2026trust}, uncertainty and mismatch~\citep{xu2026tip}, or consensus across multiple teacher traces~\citep{stein2026gates}. Objective adjustment mixes KL directions according to teacher entropy~\citep{jin2026entropy} or softens forward-KL targets where student support is low~\citep{jang2026stable}.
Signal selection and objective adjustment regulate local teacher pressure without bounding its accumulation; our ablations show that local regulation alone recovers little OOD accuracy (Table~\ref{tab:ablation}).
Drift control is well established in other learning settings: continual learning penalizes parameter or prediction changes~\citep{kirkpatrick2017overcoming,li2018learning}, RLHF anchors policies to a frozen reference via KL~\citep{ziegler2019fine,ouyang2022training}, and TRPO/PPO limit policy changes per update~\citep{schulman2015trust,schulman2017proximal}. \citet{spigler2025proximal} combines proximal policy optimization with student-driven distillation for RL control.
Our focus is on jointly regulating teacher influence and predictive drift in self-distillation: \method{} combines overlap-aware matching with a moving checkpoint penalty, jointly optimized with a K-FAC preconditioner.
\section{Conclusion}
\label{sec:conclusion}

Privileged-context self-distillation provides dense supervision but can push the student toward confident teacher predictions even where it currently assigns little probability.
\method{} regulates teacher influence and predictive drift through overlap-aware matching and a moving checkpoint penalty, both in a shared geometry and jointly optimized with a curvature preconditioner.
It retains strong ID gains while improving OOD mathematical reasoning and transfer to code.
Our analysis links forward-KL matching to rapid concentration at high-entropy states and to confident agreement on incorrect answers, whereas \method{} preserves alternatives and reduces false consensus.
These findings suggest that regulating the local influence of teacher guidance and its cumulative effect on the student's predictions can improve reasoning generalization.
We discuss limitations in Appendix~\ref{app:limitations}.

\section*{Acknowledgments}
This work is supported by the Dutch National Science Foundation (NWO Vici VI.C.212.053).

\bibliographystyle{iclr2027_conference}
\bibliography{iclr2027_conference}


\newpage
\appendix

\section{Information geometry and the natural-gradient update}
\label{app:infgeom}

This appendix lays out the geometry behind our objective and update rule. We first introduce Fisher information (\S\ref{app:fisher}) and the spherical representation of next-token distributions, which relates Hellinger divergence to Fisher--Rao distance (\S\ref{app:sphere}). We then show that these quantities and both KL directions share the same local quadratic form up to scale (\S\ref{app:local-equiv}), describe Hellinger's place in the $\alpha$-divergence family (\S\ref{app:alpha}), and examine how its gradient weights teacher preferences (\S\ref{app:influence}). Finally, we derive the local contribution of the Fisher--Rao proximal penalty to the natural-gradient update (\S\ref{app:pullback}).

\subsection{Fisher information}
\label{app:fisher}

Consider a family of distributions $p_\theta(x)$ that is smooth in $\theta$ and has support independent of $\theta$. The \emph{score} is the gradient of the log-likelihood, $\nabla_\theta\log p_\theta(x)$. Assuming differentiation and integration can be interchanged, the score has mean zero under the model
\[
    \mathbb E_{x\sim p_\theta}\!\left[\nabla_\theta\log p_\theta(x)\right]
    = \int \nabla_\theta p_\theta(x)\,dx
    = \nabla_\theta \!\int p_\theta(x)\,dx
    = 0.
\]
The \emph{Fisher information matrix} is the covariance of the score:
\begin{equation}
\label{eq:fisher-general}
    F(\theta)
    =
    \mathbb E_{x\sim p_\theta}\!\left[
        \nabla_\theta\log p_\theta(x)\,
        \nabla_\theta\log p_\theta(x)^\top
    \right]
    =
    -\,\mathbb E_{x\sim p_\theta}\!\left[
        \nabla_\theta^2\log p_\theta(x)
    \right].
\end{equation}

\paragraph{Curvature and score variability.}
The Hessian form identifies $F(\theta)$ with the expected curvature of the negative log-likelihood. For a parameter direction $v$, the quantity $v^\top F(\theta)v$ measures this curvature along $v$. The covariance form expresses the same quantity as the variance of the directional score $v^\top\nabla_\theta\log p_\theta(x)$.

\paragraph{The Fisher metric.}
Under smoothness assumptions, the KL divergence between neighboring distributions satisfies
\begin{equation}
\label{eq:fisher-kl}
    \mathrm{KL}\!\left(p_\theta\,\|\,p_{\theta+d\theta}\right)
    =
    \tfrac12\,d\theta^\top F(\theta)\,d\theta
    + O(\|d\theta\|^3),
\end{equation}
with the reverse direction having the same quadratic term (\S\ref{app:local-equiv}). Thus, parameter steps of equal Euclidean length can produce different changes in the distribution, depending on their direction. Where $F(\theta)$ is positive definite, it defines a Riemannian metric that measures local statistical distinguishability. In redundant parameterizations, including neural networks, $F(\theta)$ may be singular: directions that leave the distribution unchanged to first order have zero length under this quadratic form.

\paragraph{Invariance and the policy geometry.}
Under a smooth invertible reparameterization $\phi=\phi(\theta)$ with Jacobian $J=\partial\phi/\partial\theta$, the Fisher information transforms as $F_\theta=J^\top F_\phi J$. Thus, $d\theta^\top F_\theta d\theta=d\phi^\top F_\phi d\phi$: the local measure of distributional change is independent of the parameterization. A stronger property distinguishes the Fisher metric from other Riemannian metrics. On finite probability simplices, \v{C}encov's theorem characterizes it, up to an overall scale, by invariance under sufficient-statistic transformations \citep{amari2016information}. For next-token distributions, this metric has a closed form and a spherical representation (\S\ref{app:sphere}). Pulling it back through $\theta\mapsto\pi_\theta(\cdot\mid s)$ and averaging over states gives the policy Fisher, whose inverse defines the ideal natural-gradient preconditioner (\S\ref{app:pullback}).

\subsection{The sphere of next-token distributions}
\label{app:sphere}

A next-token distribution over a vocabulary $\mathcal{V}$ is a point in the probability simplex $\Delta_{\mathcal{V}}$. In its interior, where $p_i>0$, the Fisher metric has a simple form. For a displacement $d\mathbf p$ satisfying $\sum_i dp_i=0$, the directional score for observing category $j$ is $dp_j/p_j$. Taking its second moment under $\mathbf p$ gives
\begin{equation}
\label{eq:fisher-simplex}
    ds^2
    =
    \sum_i \frac{(dp_i)^2}{p_i},
    \qquad
    \sum_i dp_i = 0 .
\end{equation}
Thus, the metric is represented by $\mathrm{diag}(1/p_i)$ restricted to the tangent space of the simplex. It weights changes in each token's probability inversely by its current mass, so the same absolute change contributes more at a low-probability token than at a high-probability one.

\paragraph{Square-root coordinates and the spherical geometry.}
Let $u_i=\sqrt{p_i}$. Then
\[
    du_i = \frac{1}{2\sqrt{p_i}}\,dp_i,
    \qquad
    \frac{(dp_i)^2}{p_i} = 4\,(du_i)^2 ,
\]
so that
\begin{equation}
\label{eq:metric-sphere}
    ds^2 = 4\sum_i (du_i)^2 .
\end{equation}
The normalization $\sum_i p_i=1$ becomes $\sum_i u_i^2=1$. Thus, the map $\mathbf p\mapsto\sqrt{\mathbf p}$ sends the interior of the simplex onto the strictly positive orthant of the unit hypersphere $\mathbb S^{|\mathcal V|-1}$. By \eqref{eq:metric-sphere}, the Fisher metric is four times the round metric on this sphere, so the Fisher--Rao distance between two distributions is twice the length of the shortest spherical arc connecting their square-root representations.

\paragraph{Chord and arc distances.}
The square-root representations $\sqrt{\mathbf p}$ and $\sqrt{\mathbf q}$ lie on the unit hypersphere. We can measure their separation by the straight-line chord or the shortest arc along the surface. Both depend only on the overlap
\[
    \rho(\mathbf p,\mathbf q)
    :=
    \sum_i \sqrt{p_i q_i}
    =
    \cos\angle(\sqrt{\mathbf p},\sqrt{\mathbf q}),
\]
known as the Bhattacharyya coefficient.

Half the squared chord length gives the squared Hellinger divergence:
\begin{equation}
\label{eq:hellinger-chord}
    \dH(\mathbf p,\mathbf q)
    =
    \tfrac12\,\|\sqrt{\mathbf p}-\sqrt{\mathbf q}\|_2^2
    =
    1-\rho(\mathbf p,\mathbf q).
\end{equation}
The shortest arc has length $\arccos\rho(\mathbf p,\mathbf q)$ on the unit hypersphere. Accounting for the factor of $4$ in the metric, the Fisher--Rao distance is twice this length:
\begin{equation}
\label{eq:fr-arc}
    \dFR(\mathbf p,\mathbf q)
    =
    2\arccos\rho(\mathbf p,\mathbf q).
\end{equation}
Since $\rho\in[0,1]$, both quantities are bounded: $\dH\in[0,1]$ and $\dFR\in[0,\pi]$. We use the squared Hellinger divergence for distillation and the squared Fisher--Rao distance to penalize displacement from a recent checkpoint.

\subsection{Local equivalence of the divergences}
\label{app:local-equiv}

Near agreement, squared Hellinger divergence, squared Fisher--Rao distance, and both KL directions share the same quadratic term up to scale. Fix $\mathbf p$ in the interior of the simplex and write $\mathbf q=\mathbf p+\boldsymbol\epsilon$, where $\sum_i\epsilon_i=0$ and $\boldsymbol\epsilon$ is sufficiently small that $\mathbf q$ remains in the interior. Let
\[
    Q(\boldsymbol\epsilon)
    :=
    \sum_i \frac{\epsilon_i^2}{p_i}
\]
denote the Fisher quadratic form of \eqref{eq:fisher-simplex}. Expanding to second order gives
\begin{align}
    \mathrm{KL}(\mathbf p\|\mathbf q)
    &= \tfrac12\,Q(\boldsymbol\epsilon)
    + O(\|\boldsymbol\epsilon\|^3),
    \label{eq:kl-taylor}\\
    \mathrm{KL}(\mathbf q\|\mathbf p)
    &= \tfrac12\,Q(\boldsymbol\epsilon)
    + O(\|\boldsymbol\epsilon\|^3),
    \notag\\
    \dH(\mathbf p,\mathbf q)
    &= \tfrac18\,Q(\boldsymbol\epsilon)
    + O(\|\boldsymbol\epsilon\|^3),
    \label{eq:h-taylor}\\
    \dFR^2(\mathbf p,\mathbf q)
    &= Q(\boldsymbol\epsilon)
    + O(\|\boldsymbol\epsilon\|^3).
    \label{eq:fr-taylor}
\end{align}
The KL expansions follow by expanding $\log(1+\epsilon_i/p_i)$ and using $\sum_i\epsilon_i=0$. For Hellinger and Fisher--Rao, expanding
\[
    \sqrt{p_iq_i}
    =
    p_i+\tfrac12\epsilon_i
    -\tfrac18\frac{\epsilon_i^2}{p_i}
    +O(|\epsilon_i|^3)
\]
and summing over tokens gives
\[
    \rho
    =
    1-\tfrac18 Q(\boldsymbol\epsilon)
    +O(\|\boldsymbol\epsilon\|^3).
\]
Substituting into \eqref{eq:hellinger-chord} yields \eqref{eq:h-taylor}. Using
$[\arccos(1-\delta)]^2=2\delta+O(\delta^2)$
in \eqref{eq:fr-arc} yields \eqref{eq:fr-taylor}. Moreover, since $\delta=1-\rho=\dH(\mathbf p,\mathbf q)$ is of second order in $\boldsymbol\epsilon$,
\[
    \dFR^2(\mathbf p,\mathbf q)
    =
    8\,\dH(\mathbf p,\mathbf q)
    +
    O(\|\boldsymbol\epsilon\|^4).
\]
After accounting for scale, the squared chord and arc distances agree through third order and first differ at fourth order.

These expansions show that the losses have the same local geometry up to scale. Their differences become relevant away from agreement, where the quadratic approximation may no longer describe them well. This motivates examining their behavior at the low-overlap states encountered in privileged-context OPSD, where the teacher can assign high probability to tokens the student barely supports. The following sections compare the divergences and their gradients in this regime.

\subsection{Hellinger and the two KL directions}
\label{app:alpha}

\paragraph{The two KL directions.}
Distillation can match the teacher $q$ and student $p_\theta$ in either KL direction. The \emph{forward} direction,
\begin{equation}
\label{eq:forward-kl}
\mathrm{KL}(q\|p_\theta)
=
\sum_i q_i\log\frac{q_i}{p_\theta(i)},
\end{equation}
diverges if the student assigns zero probability to any token with positive teacher probability. It therefore penalizes missing teacher-supported mass, a property often described as mass-covering or zero-avoiding. The \emph{reverse} direction,
\begin{equation}
\label{eq:reverse-kl}
\mathrm{KL}(p_\theta\|q)
=
\sum_i p_\theta(i)\log\frac{p_\theta(i)}{q_i},
\end{equation}
instead diverges if the student assigns positive probability to a token with zero teacher probability. Because its terms are weighted by student mass, it places less pressure on teacher-supported regions to which the student assigns little probability. When the student cannot represent the teacher distribution exactly, this asymmetry can favor concentrating on a subset of teacher modes, a tendency known as mode-seeking. These descriptions characterize tendencies under approximation constraints; both KL directions are minimized at $p_\theta=q$ when an exact match is possible.

\paragraph{Hellinger within the $\alpha$-divergence family.}
The two KL directions arise as limits of the $\alpha$-divergences,
\begin{equation}
\label{eq:alpha-divergence}
D_\alpha(\mathbf p\|\mathbf q)
=
\frac{4}{1-\alpha^2}
\left(
1-\sum_i p_i^{(1-\alpha)/2}\,q_i^{(1+\alpha)/2}
\right),
\qquad
\alpha\in(-1,1).
\end{equation}
For strictly positive distributions, the limits are
\[
    \lim_{\alpha\to 1}D_\alpha(\mathbf p\|\mathbf q)
    =\mathrm{KL}(\mathbf q\|\mathbf p),
    \qquad
    \lim_{\alpha\to -1}D_\alpha(\mathbf p\|\mathbf q)
    =\mathrm{KL}(\mathbf p\|\mathbf q).
\]
At $\alpha=0$, the family recovers squared Hellinger divergence up to scale:
\begin{equation}
D_0(\mathbf p\|\mathbf q)
=
4\left(1-\sum_i\sqrt{p_iq_i}\right)
=
4\,\dH(\mathbf p,\mathbf q).
\end{equation}
The family satisfies
$D_\alpha(\mathbf p\|\mathbf q)=D_{-\alpha}(\mathbf q\|\mathbf p)$,
so $\alpha=0$ is its self-dual member: exchanging the student and teacher leaves the divergence unchanged. Hellinger is therefore the symmetric midpoint in this parameterization. Its implications for teacher influence follow from the gradient, which we examine next.

\subsection{Per-token teacher influence}
\label{app:influence}

The Hellinger gradient makes its dependence on student--teacher overlap explicit. Differentiating the Hellinger term in \eqref{eq:hellinger-chord} while holding the teacher distribution $q$ fixed gives the form used in \eqref{eq:hellinger-grad},
\begin{equation}
\label{eq:hellinger-grad-app}
    \nabla_\theta \dH(p_\theta,q)
    =
    -\tfrac12\sum_{i\in\mathcal{V}}
    \sqrt{p_\theta(i)\,q_i}\;
    \nabla_\theta\log p_\theta(i),
\end{equation}
using $\nabla_\theta\sqrt{p_\theta(i)}
=\tfrac12\sqrt{p_\theta(i)}\,\nabla_\theta\log p_\theta(i)$.
The three matching losses considered here can all be written in weighted-score form:
\[
    \nabla_\theta \mathcal{L}
    =
    -\sum_{i\in\mathcal{V}} w_i\,\nabla_\theta\log p_\theta(i).
\]
Their weights are
\begin{align}
    \text{Forward KL }\ \mathrm{KL}(q\|p_\theta):\quad
        & w_i = q_i, \label{eq:w-fkl}\\
    \text{Reverse KL }\ \mathrm{KL}(p_\theta\|q):\quad
        & w_i = p_\theta(i)\!\left(
            \log\tfrac{q_i}{p_\theta(i)}
            +\mathrm{KL}(p_\theta\|q)
        \right),
        \label{eq:w-rkl}\\
    \text{Hellinger }\ \dH(p_\theta,q):\quad
        & w_i = \tfrac12\sqrt{p_\theta(i)\,q_i}.
        \label{eq:w-h}
\end{align}
The term $\mathrm{KL}(p_\theta\|q)$ in \eqref{eq:w-rkl} centers the weights so that $\sum_i w_i=0$. Adding any token-independent scalar to the log-ratio term leaves the weighted-score sum unchanged, since
$\sum_i p_\theta(i)\nabla_\theta\log p_\theta(i)=0$.
With the stated baseline, $w_i<0$ whenever
$\log(p_\theta(i)/q_i)>\mathrm{KL}(p_\theta\|q)$.

A positive weight contributes an update in the direction of increasing $\log p_\theta(i)$, but the net effect also depends on the other tokens. For example, if $p_\theta=\mathrm{softmax}(z)$, then
\[
    \frac{\partial\mathcal L}{\partial z_i}
    =
    p_\theta(i)\sum_j w_j-w_i.
\]
For Hellinger, this becomes
\[
    \frac{\partial\dH(p_\theta,q)}{\partial z_i}
    =
    \tfrac12\left(
        \rho(p_\theta,q)\,p_\theta(i)
        -\sqrt{p_\theta(i)\,q_i}
    \right).
\]

The weights differ in how they treat teacher preferences with little student support. Forward KL assigns weight $q_i$ regardless of $p_\theta(i)$, so a teacher-preferred token retains its weight even as the student's probability approaches zero. Reverse KL weights the log-ratio signal by student probability; for a fixed strictly positive teacher distribution, this weight vanishes as $p_\theta(i)\to0$. Hellinger's weight also vanishes in this limit, but scales as the geometric mean $\sqrt{p_\theta(i)q_i}$. For fixed teacher probability, it increases with the square root of student probability, allowing greater influence as the student assigns more mass to that token.

These expressions explain how Hellinger attenuates teacher preferences with little student support, as illustrated in Figure~\ref{fig:overview}A. They do not, by themselves, imply that the parameter-gradient norm grows monotonically as overlap falls for forward KL, or establish a parameter-independent bound for Hellinger: the score vectors also determine the norm. Figure~\ref{fig:mechanism}A shows the empirical relationship.

\subsection{Natural-gradient update and the checkpoint pullback}
\label{app:pullback}

Both terms of the objective compare predictive distributions using the geometry of \S\ref{app:sphere}. This also motivates using the Fisher metric to scale the update: parameter steps of equal Euclidean length can produce very different changes in the predictions.

\paragraph{The proximal term has a local Fisher form.}
We analyze the proximal penalty with the sampled trajectories held fixed, consistent with the training objective. Let
\[
    \Delta_k := \theta_k-\theta_{\mathrm{ckpt}},
\]
and write $F(\theta;s)$ for the Fisher information of the next-token distribution $\pi_\theta(\cdot\mid s)$ at a fixed state $s=(x,y_{<t})$. A second-order expansion around the checkpoint gives
\[
    \dFR^2\!\left(
        \pi_{\theta_k}(\cdot\mid s),
        \pi_{\theta_{\mathrm{ckpt}}}(\cdot\mid s)
    \right)
    =
    \Delta_k^\top F(\theta_{\mathrm{ckpt}};s)\Delta_k
    + O(\|\Delta_k\|^3).
\]
Let $F_{\mathrm{ckpt}}$ denote the average of
$F(\theta_{\mathrm{ckpt}};s)$ over these states, using the same weighting as the proximal loss. Averaging the expansion gives
\begin{equation}
\label{eq:prox-local}
    \Lprox(\theta_k;\theta_{\mathrm{ckpt}})
    \approx
    \Delta_k^\top F_{\mathrm{ckpt}}\Delta_k.
\end{equation}
Thus, locally,
\[
    \nabla_\theta \Lprox(\theta_k;\theta_{\mathrm{ckpt}})
    \approx
    2\,F_{\mathrm{ckpt}}\Delta_k.
\]
The matrix $F_{\mathrm{ckpt}}$ appears only in this local analysis; the algorithm does not need to construct it. Training evaluates the proximal penalty directly from the student and checkpoint distributions.

\paragraph{The ideal natural-gradient contribution.}
Let $F_k$ denote the current policy Fisher, evaluated at $\theta_k$ and averaged over the same states. Assuming $F_k$ is invertible, applying the natural-gradient preconditioner to the local proximal gradient gives
\begin{equation}
\label{eq:app-checkpoint-pullback}
    -\eta\lambda\,
    F_k^{-1}\nabla_\theta\Lprox(\theta_k;\theta_{\mathrm{ckpt}})
    \approx
    -2\eta\lambda\,
    F_k^{-1}F_{\mathrm{ckpt}}\Delta_k.
\end{equation}
Here, $F_{\mathrm{ckpt}}$ weights parameter displacement by its effect on the checkpoint's predictions, while $F_k^{-1}$ scales the update according to the current policy's geometry.

If $F_k^{-1}F_{\mathrm{ckpt}}\approx I$, the proximal contribution is approximately
\[
    -2\eta\lambda(\theta_k-\theta_{\mathrm{ckpt}}).
\]
This has the same form as a Euclidean gradient step on
$\lambda\lVert\theta-\theta_{\mathrm{ckpt}}\rVert_2^2$.
The equivalence is about the local update, not the penalty itself. The Fisher--Rao penalty still compares predictive distributions. If the Fisher is singular, a pseudoinverse restricts this interpretation to the identifiable parameter directions.

\paragraph{The implemented preconditioner.}
In practice, we use a damped, block-diagonal K-FAC construction described in Appendix~\ref{app:exp-kfac}. Writing the implemented preconditioning operator as $P_k$, its local proximal contribution is
\[
    -\eta\lambda P_k
    \nabla_\theta\Lprox(\theta_k;\theta_{\mathrm{ckpt}})
    \approx
    -2\eta\lambda P_kF_{\mathrm{ckpt}}\Delta_k.
\]
\section{Experimental details}
\label{app:exp-details}

\subsection{Models}
\label{app:exp-models}
We evaluate along two axes: scale (Qwen3 dense at 1.7B/4B/8B/14B/32B) and
cross-family (Qwen3-8B, Olmo-3-7B-Think, DeepSeek-R1-Distill-Llama-8B), taking each
released checkpoint as the base model $\pi_0$. Architectures are listed in Table~\ref{tab:archs}.

\subsection{Data and privileged context}
\label{app:exp-data}
We sample $1$K problems from the English-filtered portion of DAPO-Math, which provides verifiable integer answers but no reference solutions. We construct the privileged-solution pool offline by sampling up to $32$ solutions per problem from $\pi_0$ at temperature $1.0$ and retaining the rollouts whose final answers match the reference answer. Problems with no verified solution are excluded from training. The resulting pool contains $7$K verified solutions. During training, we draw the privileged context $c$ from the verified solutions for the current problem, $c\sim\mathcal C(\cdot\mid x)$.

\subsection{Training}
\label{app:exp-rollouts}

Each distillation update uses a batch of $8$ problems. The student samples one rollout per problem from the question alone. The teacher uses the same model parameters, receives a sampled verified solution as additional context, and scores the student's trajectory at each response position. GRPO instead samples $G=8$ rollouts for one problem per update. This matches the number of generated completions per update, although the number of prompts differs. For both distillation and GRPO, rollouts are sampled from the current student at each step, using temperature $1.0$ and a maximum response length of $4096$ tokens.
We train for $625$ optimizer steps. For \method{}, we refresh the proximal checkpoint $\theta_{\mathrm{ckpt}}$ every $K_{\mathrm{ckpt}}=64$ steps, allowing the reference distribution to track the student's recent predictions.
We use top-$K$ truncation with $K=1024$ when computing the divergence losses. For \method{}, the retained token set at each position is the union of the student, teacher, and checkpoint top-$K$ sets. This set contains $>99.9\%$ of the teacher's probability mass at the loss-driving states, measured before renormalization. We then renormalize each distribution over the shared set by applying a restricted softmax.

\subsection{Evaluation}
\label{app:exp-eval}

We evaluate in-distribution performance on a held-out set of $500$ DAPO-Math problems and out-of-distribution mathematical reasoning on AIME~2024, AIME~2025, AMC~2023, and MATH-500. We sample $16$ responses per problem at temperature $0.6$ and top-$p=0.95$, with a maximum response length of $8192$ tokens. We report avg@$16$, the average accuracy across the sampled responses, and pass@$16$, the fraction of problems with at least one correct response. Aggregate OOD mathematics accuracy is the unweighted mean across the four benchmarks.

To assess cross-domain transfer, we evaluate Qwen3-8B on HumanEval+, MBPP+, and LiveCodeBench after training exclusively on DAPO-Math. We sample $16$ responses per problem at temperature $0.6$ and top-$p=0.95$, with a maximum response length of $8192$ tokens. Aggregate OOD code accuracy is the unweighted mean across these three benchmarks.

\begin{table}
\centering\small
\caption{Model architectures. $L$: blocks; $d$: hidden size;
$d_{\mathrm{ff}}$: FFN width; heads/kv: query heads/key--value heads; $h$: head dim; $V$: vocabulary.}
\label{tab:archs}
\begin{tabular}{lcccccc}
\toprule
Model & $L$ & $d$ & $d_{\mathrm{ff}}$ & heads/kv & $h$ & $V$ \\
\midrule
Qwen3-1.7B             & 28 & 2048 & 6144  & 16/8  & 128 & $\approx$152K \\
Qwen3-4B               & 36 & 2560 & 9728  & 32/8  & 128 & $\approx$152K \\
Qwen3-8B               & 36 & 4096 & 12288 & 32/8  & 128 & $\approx$152K \\
Qwen3-14B              & 40 & 5120 & 17408 & 40/8  & 128 & $\approx$152K \\
Qwen3-32B              & 64 & 5120 & 25600 & 64/8  & 128 & $\approx$152K \\
Olmo-3-7B-Think        & 32 & 4096 & 11008 & 32/32 & 128 & $\approx$100K \\
DeepSeek-R1-Distill-Llama-8B & 32 & 4096 & 14336 & 32/8  & 128 & $\approx$128K \\
\bottomrule
\end{tabular}
\end{table}

\begin{table}
\centering\small
\caption{Hyperparameters. Rollout and evaluation rows are identical across methods.}
\label{tab:hparams}
\begin{tabular}{ll}
\toprule
Hyperparameter & Value \\
\midrule
Optimizer (baselines)     & AdamW ($\beta_1{=}0.9,\ \beta_2{=}0.95$), lr $1\times10^{-6}$, 20-step warmup \\
Optimizer (\method{})     & K-FAC, lr $1\times10^{-6}$ (jointly swept with $\gamma$), 20-step warmup \\
Optimizer steps           & 625 \\
Precision                 & bf16 compute, fp32 master weights \\
Generations / update      & 8 (matched); 1 rollout/prompt (distillation), $G{=}8$ (GRPO) \\
Training cap / eval cap   & 4096 / 8192 tokens \\
Solution construction     & up to 32 rollouts/problem \\
Decoding (eval)           & temperature 0.6, top-$p$ 0.95; avg@16, pass@16 \\
\method{}: $\lambda$ / checkpoint refresh / top-$K$ & 1.0 / 64 steps / 1024 \\
\method{}: K-FAC $g$ / $\gamma$ / subsample $s$ & 16 ($\le$8B), 32 (14B,32B) / $10^{-3}$ / 1/8 \\
\bottomrule
\end{tabular}
\end{table}

\subsection{Baselines}
\label{app:exp-baselines}

All on-policy distillation methods use a solution-conditioned teacher, student-generated prefixes, $8$ rollouts per update, $625$ optimizer steps, and top-$K$ logit truncation. GRPO uses the same number of rollouts per update and optimizer steps, with the different prompt allocation described in \S\ref{app:exp-rollouts}. SFT trains on offline solution traces, and the base model receives no updates. We use AdamW for the trained baselines and the K-FAC preconditioner described in \S\ref{app:exp-kfac} for \method{}. Optimizer settings are listed in Table~\ref{tab:hparams}.

For hyperparameter selection, we train each candidate configuration on a fixed set of $100$ DAPO-Math problems and select configurations by validation avg@$8$ on a separate set of $200$ problems, using $8$ responses per validation problem. The tuning and validation sets are mutually disjoint and do not overlap with the main training set or the final ID test set. For the AdamW baselines, we sweep the learning rate over $\{10^{-7},\,3\times10^{-7},\,10^{-6},\,3\times10^{-6},\,10^{-5}\}$ and select $10^{-6}$.

\paragraph{Reference baselines.}
Base ($\pi_0$) is the released checkpoint without post-training. SFT minimizes cross-entropy on the verified solution traces. GRPO \citep{shao2024deepseekmath} uses a binary reward for final-answer correctness, a group size of $8$, and no privileged context. We sweep its KL coefficient over $\{0,0.01,0.03,0.1\}$ and clipping parameter over $\{0.1,0.2\}$, selecting $\beta_{\mathrm{KL}}=0.03$ and $\epsilon=0.2$.

\paragraph{Divergence controls.}
FwdKL \citep{zhao2026selfdistilled} minimizes $\mathrm{KL}(q\|p_\theta)$, while RevKL \citep{gu2024minillm} minimizes $\mathrm{KL}(p_\theta\|q)$. Both use a stop-gradient teacher and have no additional tuned hyperparameters beyond the shared training settings. For JSD \citep{agarwal2024gkd}, we sweep $\beta\in\{0.1,0.25,0.5,0.75,0.9\}$ and select $\beta=0.5$. For SkewKL \citep{ko2024distillm,ko2025distillm}, we sweep $\alpha\in\{0.01,0.03,0.1,0.3,0.5\}$ and select $\alpha=0.1$.

\paragraph{Filtering baselines.}
\emph{TrOPD} \citep{xing2026trust}.
We independently assign each sampled position to the trust region with probability $\min\!\big(q^{(t)}(y_t)/p^{(t)}_\theta(y_t),\,1\big)$. Trust-region positions use the reverse-KL $K_1$ estimator $\log\!\big(p^{(t)}_\theta(y_t)/q^{(t)}(y_t)\big)$; outliers use forward KL restricted to the teacher's top-$k$ tokens. An additional forward-KL term, weighted by $\beta$, uses prefixes of the sampled verified solution in place of the original method's teacher-generated prefixes, with prefix length cosine-annealed to zero. We sweep $k \in \{16, 64, 256, 1024\}$ and $\beta \in \{0, 0.001, 0.01, 0.1\}$, selecting $k = 64$ and $\beta = 0.001$.

\emph{TIP} \citep{xu2026tip}.
We retain the top-$r$ fraction of positions in each batch according to the Soft-OR score $s_t = \hat h_t + \hat\delta_t - \hat h_t\hat\delta_t$. Here, $\hat h_t$ and $\hat\delta_t$ denote student entropy and per-position reverse KL $\mathrm{KL}(p^{(t)}_\theta\,\|\,q^{(t)})$, respectively, after each quantity is clipped at its $p$-quantile within the batch and min--max normalized to $[0,1]$. The loss averages the reverse KL over the retained positions. We sweep $r \in \{0.1, 0.2, 0.5, 1.0\}$ and $p \in \{0.90, 0.95, 0.98, 0.99, 1.00\}$, selecting $r = 0.5$ and $p = 0.98$.

\paragraph{\method{}.}
We tune the proximal coefficient $\lambda$, checkpoint-refresh interval $K_{\mathrm{ckpt}}$, learning rate $\eta$, and damping parameter $\gamma$. Because damping affects the scale of the preconditioned update, we tune the learning rate and damping jointly:
\[
    (\eta,\gamma)
    \in
    \{10^{-7},10^{-6},10^{-5}\}
    \times
    \{10^{-4},10^{-3}\}.
\]
We also consider $\lambda\in\{0.1,1.0,3.0\}$ and
$K_{\mathrm{ckpt}}\in\{32,64\}$.
The selected configuration is $\eta=10^{-6}$, $\gamma=10^{-3}$,
$\lambda=1.0$, and $K_{\mathrm{ckpt}}=64$.

\paragraph{Concurrent work.}
Several concurrent studies explore related ways to regulate OPD. For distillation from an external teacher, \citet{yu2026mismatch} handle degenerate agreement by combining bounded Hellinger shaping for student-excess tokens with a teacher top-$K$ injection that restores probability mass on underweighted teacher-preferred tokens. \citet{xie2026trust} interpolate the student and teacher distributions into a proximal teacher, then use trust-region updates to train the student. \citet{plyusov2026trust} change the rollout policy during warmup, moving it toward the teacher within a student-centered KL trust region but keep the distillation loss fixed. In privileged-context OPSD, \citet{liu2026self} combine verifier-based policy gradients with full-vocabulary reverse-KL matching and KL regularization to a fixed reference policy. We instead study OOD generalization under privileged-context self-distillation, using overlap-aware matching and a moving checkpoint penalty.
Unlike \citet{yu2026mismatch}, who apply Hellinger shaping to selected sampled tokens under a fixed external teacher, we match full next-token distributions in a setting where the teacher shares the student's parameters, so overlap weighting alone slows but does not stop drift from compounding (Figure~\ref{fig:mechanism}B). The gains depend on combining Hellinger matching with the checkpoint penalty. With K-FAC held fixed, Hellinger matching alone reaches $53.8$ OOD accuracy, and the penalty with forward KL reaches only $50.6$, below the base model's $51.9$, whereas together they reach $60.5$ (Table~\ref{tab:ablation}).

\subsection{Kronecker factorization}
\label{app:exp-kfac}

\method{} uses a preconditioner based on K-FAC \citep{martens2015kfac}. We construct its factors from activations and pre-activation gradients obtained by backpropagating the training loss on the on-policy batch. This reuses the existing backward pass. 
For a linear layer $W\in\mathbb R^{d_{\rm out}\times d_{\rm in}}$, let $a$ denote the input activation and $\delta$ the gradient of the training loss with respect to the layer's pre-activation.
We form the second-moment factors
\[
    A=\hat{\mathbb E}[aa^\top]
    \in\mathbb R^{d_{\rm in}\times d_{\rm in}},
    \qquad
    G=\hat{\mathbb E}[\delta\delta^\top]
    \in\mathbb R^{d_{\rm out}\times d_{\rm out}},
\]
where $\hat{\mathbb E}$ averages over sampled token positions in the training batch. With column-wise vectorization of $W$, the Kronecker-factored matrix is $A\otimes G$.

We maintain exponential moving averages of these factors and apply factored damping. Using $A$ and $G$ to denote the block-diagonal running estimates, the preconditioned gradient is
\[
    \widetilde{\nabla_W\mathcal L}
    =
    (G+\sqrt{\gamma}\,I)^{-1}
    (\nabla_W\mathcal L)
    (A+\sqrt{\gamma}\,I)^{-1}.
\]
This follows from the Kronecker inverse identity and requires solving with the two factors rather than the full matrix. We apply this preconditioner to the seven projections ($Q,K,V,O,\mathrm{gate},\mathrm{up},\mathrm{down}$) in each transformer block. Factor statistics are accumulated on a $1/8$ subsample of token positions, with $\gamma=10^{-3}$. We use block-diagonal factors stored in fp32 with $g=16$ for models up to $8$B, and in bf16 with $g=32$ for the $14$B and $32$B models. The memory requirements are reported in Appendix~\ref{app:compute}, Table~\ref{tab:kfac_memory}.

\subsection{Hardware}
\label{app:exp-hardware}
We run the main experiments on NVIDIA H200 GPUs, with rollout generation and checkpoint scoring placed on inference workers and optimization on training workers. All methods in a given comparison use the same hardware class and precision.

\section{Additional results}
\label{app:results}

\subsection{The ID--OOD trade-off}
\label{app:frontier}

\begin{figure}
    \centering
    \includegraphics[]{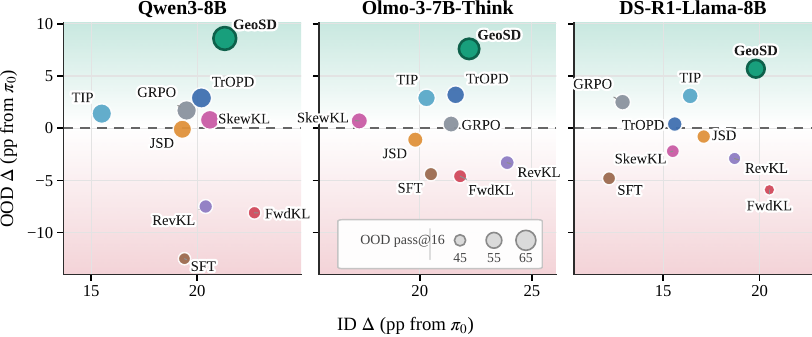}
    \caption{In-distribution vs.\ out-of-distribution gains (avg@$16$ percentage points relative
    to $\pi_0$) for every method across the three model families; dot size encodes OOD
    pass@$16$.}
    \label{fig:frontier}
\end{figure}

Figure~\ref{fig:frontier} summarizes the ID--OOD trade-off across the three model families. \method{} achieves the highest average OOD accuracy while retaining strong ID gains, whereas forward and reverse KL improve ID accuracy at the expense of OOD performance.

\begin{figure}
    \centering
    \includegraphics[]{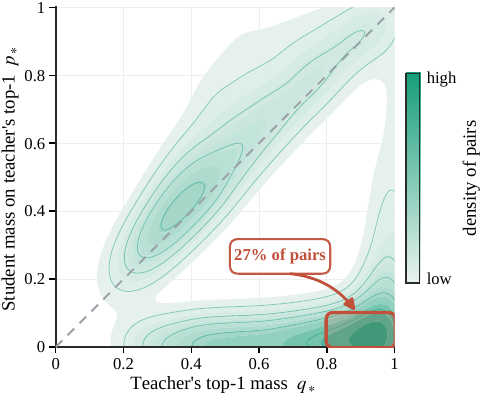}
    \caption{
    Teacher top-token support at high-entropy positions (top $10$\% by $\pi_0$ entropy).
    We plot the teacher's top-1 mass against the student's mass on the same token. Density in the lower-right region identifies states where privileged context makes the teacher confident in a continuation that the student assigns little probability.
    }
    \label{fig:overlap_density}
\end{figure}

\subsection{Teacher top-token support}
\label{app:teacher-top-token-support}

To characterize teacher--student mismatch at initialization, we use Qwen3-8B and collect the top $10\%$ of response positions by $\pi_0$ next-token entropy (as in \S\ref{sec:high-ent}) from the initial on-policy training batch, before any parameter updates. The teacher receives a full answer-verified solution as privileged context and scores the same prefixes as the student. We measure probabilities over the full vocabulary.

At each position, we identify the teacher's most probable token, $j_\star=\arg\max_i q_i$, and record its probability under the teacher, $q_\star=q_{j_\star}$, and student, $p_\star=p_{j_\star}$. Figure~\ref{fig:overlap_density} shows the joint density of these probabilities. The dashed diagonal marks equal teacher and student probability on the selected token. Positions satisfying $q_\star\ge0.8$ and $p_\star\le0.1$ account for $27\%$ of these high-entropy positions. At these states, the privileged teacher strongly favors a token to which the student assigns little probability, creating the mismatch that overlap-aware matching is designed to attenuate.

\subsection{Per-benchmark OOD generalization}
\label{app:results-per-bench}

Table~\ref{tab:ood_per_bench} breaks down the aggregate OOD accuracy in Table~\ref{tab:main_id_ood} by benchmark. \method{} achieves the highest mean accuracy on all four benchmarks across all three model families, improving over $\pi_0$ in every case. The largest gains over $\pi_0$ occur on MATH-500 for Qwen3-8B and on AIME~2025 for both Olmo-3-7B-Think and DS-R1-Llama-8B. In contrast, RevKL falls below $\pi_0$ on every mathematics benchmark in all three families, while FwdKL does so in all but one case: on MATH-500 with Olmo-3-7B-Think, it improves by 0.2 percentage points. Both KL objectives reduce aggregate OOD accuracy in every family.

Table~\ref{tab:code} reports the per-benchmark code results for Qwen3-8B after training exclusively on DAPO-Math. \method{} achieves the highest mean accuracy on all three benchmarks, improving over $\pi_0$ by $2.4$ percentage points on HumanEval+, $1.9$ on MBPP+, and $1.4$ on LiveCodeBench. The average improvement is $1.9$ percentage points. SFT, FwdKL, and RevKL fall below $\pi_0$ on every code benchmark, while GRPO and the filtering baselines produce smaller or mixed gains. These results provide evidence of positive transfer from mathematics to code generation without code supervision during post-training.

\begin{table}[t]
\centering
\caption{Per-benchmark OOD performance (avg@16) under the full-solution teacher, for the three model families.}
\label{tab:ood_per_bench}
\resizebox{\textwidth}{!}{%
\begin{tabular}{l cccc cccc cccc}
\toprule
& \multicolumn{4}{c}{\textbf{Qwen3-8B}}
& \multicolumn{4}{c}{\textbf{Olmo-3-7B-Think}}
& \multicolumn{4}{c}{\textbf{DS-R1-Llama-8B}} \\
\cmidrule(lr){2-5}
\cmidrule(lr){6-9}
\cmidrule(lr){10-13}
\textbf{Method}
& AIME24 & AIME25 & AMC23 & MATH-500
& AIME24 & AIME25 & AMC23 & MATH-500
& AIME24 & AIME25 & AMC23 & MATH-500 \\
\midrule
\multicolumn{13}{l}{\emph{Reference baselines}}\\
$\pi_0$
& 32.1 & 31.6 & 65.1 & 78.8
& 29.8 & 24.0 & 62.3 & 72.7
& 27.1 & 19.6 & 56.3 & 72.0 \\
SFT
& 19.0 & 19.7 & 53.3 & 65.6
& 23.5 & 19.7 & 57.4 & 70.7
& 22.5 & 17.4 & 52.4 & 63.3 \\
GRPO
& 32.5 & 33.1 & 69.0 & 79.6
& 29.7 & 26.5 & 61.2 & 72.9
& 30.6 & 24.5 & 57.6 & 72.2 \\
\midrule
\multicolumn{13}{l}{\emph{Divergence controls}}\\
FwdKL
& 22.4 & 22.9 & 58.6 & 71.2
& 22.5 & 17.8 & 57.3 & 72.9
& 21.3 & 16.7 & 50.2 & 63.0 \\
RevKL
& 23.8 & 22.7 & 60.3 & 71.0
& 26.0 & 19.0 & 59.4 & 71.3
& 24.1 & 17.0 & 53.6 & 68.6 \\
JSD
& 29.0 & 31.0 & 69.2 & 77.9
& 29.3 & 23.0 & 59.3 & 72.8
& 27.3 & 19.1 & 55.4 & 69.8 \\
SkewKL
& 30.4 & 33.4 & 66.3 & 80.7
& 29.9 & 22.8 & 63.3 & 75.6
& 23.8 & 18.7 & 57.3 & 66.2 \\
\midrule
\multicolumn{13}{l}{\emph{Filtering baselines}}\\
TrOPD
& 33.7 & 35.9 & 68.4 & 81.4
& 31.7 & 24.6 & 66.2 & 79.2
& 27.1 & 21.2 & 56.3 & 71.8 \\
TIP
& 31.6 & 33.7 & 69.3 & 78.7
& 29.9 & 26.2 & 65.4 & 78.8
& 30.4 & 25.5 & 60.3 & 71.0 \\
\midrule
\method{}
& 39.5 & 38.6 & 74.6 & 89.3
& 33.7 & 34.0 & 69.0 & 82.4
& 33.0 & 26.6 & 62.4 & 75.7 \\
\bottomrule
\end{tabular}%
}
\end{table}
\begin{table}[t]

\caption{Code reasoning under cross-domain OOD transfer
(Qwen3-8B, avg@16).
Means over ten runs, with sample
standard deviations in subscripts.
Training uses DAPO-Math only, with no code supervision during
post-training; HumanEval+ (HE+), MBPP+, and LiveCodeBench (LCB)
are evaluated zero-shot.
OOD avg.\ is the unweighted mean across the three benchmarks.
$\Delta_{\mathrm{OOD}}$ is relative to $\pi_0$.
Best measured result in bold; second-best underlined.}
\label{tab:code}
\centering
\resizebox{\textwidth}{!}{%
\begin{tabular}{l c cc cccc cc c}
\toprule
& & \multicolumn{2}{c}{\textit{Reference}}
& \multicolumn{4}{c}{\textit{Divergence}}
& \multicolumn{2}{c}{\textit{Filtering}} & \\
\cmidrule(lr){3-4}
\cmidrule(lr){5-8}
\cmidrule(lr){9-10}
& $\pi_0$ & SFT & GRPO
& FwdKL & RevKL & JSD & SkewKL
& TrOPD & TIP & \textsc{\method{}} \\
\midrule
HE+
& $62.1_{0.7}$ & $56.5_{1.7}$ & $62.9_{1.1}$
& $57.5_{1.6}$ & $59.5_{2.1}$
& $62.0_{1.1}$ & $62.5_{1.3}$
& $\underline{63.2}_{1.6}$ & $61.4_{1.1}$
& $\mathbf{64.5}_{0.4}$ \\
MBPP+
& $56.4_{0.5}$ & $51.5_{1.2}$ & $57.0_{1.2}$
& $52.5_{1.1}$ & $54.0_{1.5}$
& $56.3_{0.8}$ & $56.8_{1.0}$
& $\underline{57.4}_{1.3}$ & $57.2_{0.6}$
& $\mathbf{58.3}_{0.6}$ \\
LCB
& $21.4_{0.4}$ & $17.5_{0.7}$ & $\underline{22.1}_{0.8}$
& $18.0_{0.7}$ & $19.5_{1.1}$
& $21.3_{0.6}$ & $21.6_{0.7}$
& $20.9_{0.9}$ & $21.9_{0.6}$
& $\mathbf{22.8}_{0.3}$ \\
\addlinespace[2pt]
OOD avg.
& $46.6_{0.4}$ & $41.8_{0.7}$ & $\underline{47.3}_{0.4}$
& $42.7_{0.7}$ & $44.3_{0.9}$
& $46.5_{0.5}$ & $47.0_{0.6}$
& $47.2_{0.8}$ & $46.8_{0.3}$
& $\mathbf{48.5}_{0.2}$ \\
$\Delta_{\mathrm{OOD}}$
& -- & $-4.8$ & $\underline{+0.7}$
& $-3.9$ & $-2.3$
& $-0.1$ & $+0.4$
& $+0.6$ & $+0.2$
& $\mathbf{+1.9}$ \\
\bottomrule
\end{tabular}%
}
\end{table}

\subsection{Statistical analysis of OOD gains}
We assess differences in aggregate OOD accuracy using the run as the unit of analysis ($n=10$ per method). For each run, we compute the unweighted mean of avg@$16$ across the four mathematics benchmarks. Within each model family, we compare \method{} with the baseline having the highest mean aggregate OOD accuracy, using a one-sided Welch's $t$-test with the alternative hypothesis that \method{} has higher expected accuracy.
We control the family-wise error rate across the three families with the Holm correction at $\alpha=0.05$ and report the mean gain $\Delta$, its 95\% CI, and Hedges' $g$. \method{} significantly outperforms the best baseline in every family (Table~\ref{tab:sig}; all Holm-adjusted $p<10^{-5}$, $g\geq 3.1$).

\begin{table}
\centering
\caption{Significance test on aggregate OOD accuracy (avg@16, mean over the four benchmarks). \method{} vs.\ the strongest baseline per family, one-sided Welch's $t$-test ($n=10$ runs), Holm-corrected across families. $\Delta$: mean gain over the best baseline; $g$: Hedges' effect size.}
\label{tab:sig}
\begin{tabular}{lccccc}
\toprule
Family & \method{} & Best baseline & $\Delta$ (95\% CI) & Hedges' $g$ & $p$ (Holm) \\
\midrule
Qwen3-8B        & 60.5 & TrOPD (54.8) & $+5.7\ [4.9,\,6.5]$ & 6.5 & $2.9\times10^{-9}$ \\
Olmo-3-7B-Think & 54.8 & TrOPD (50.4) & $+4.4\ [3.8,\,5.0]$ & 6.7 & $1.1\times10^{-11}$ \\
DS-R1-Llama-8B  & 49.4 & TIP (46.8)   & $+2.6\ [1.8,\,3.4]$ & 3.1 & $1.3\times10^{-6}$ \\
\bottomrule
\end{tabular}
\end{table}

\subsection{Robustness to step size and early stopping}
\label{app:stepsize}

Figure~\ref{fig:mechanism}A shows that Hellinger and forward-KL gradients differ in both shape and typical magnitude. We therefore examine whether smaller learning rates or earlier stopping allow FwdKL to recover the ID and OOD improvements of \method{}. We sweep the learning rate over $\{3\times10^{-8},\,10^{-7},\,3\times10^{-7},\,10^{-6},\,3\times10^{-6},\,10^{-5}\}$ for FwdKL and over $\{10^{-7},\,3\times10^{-7},\,10^{-6},\,3\times10^{-6},\,10^{-5}\}$ for \method{}. Each configuration is trained for $625$ updates, with all other settings fixed to the corresponding method's defaults in Table~\ref{tab:hparams}. We evaluate every $25$ updates through update $125$ and every $125$ updates thereafter. This comparison evaluates the two complete methods, including their respective optimizers and regularization settings.

\begin{figure}[t]
\centering
\includegraphics[width=\linewidth]{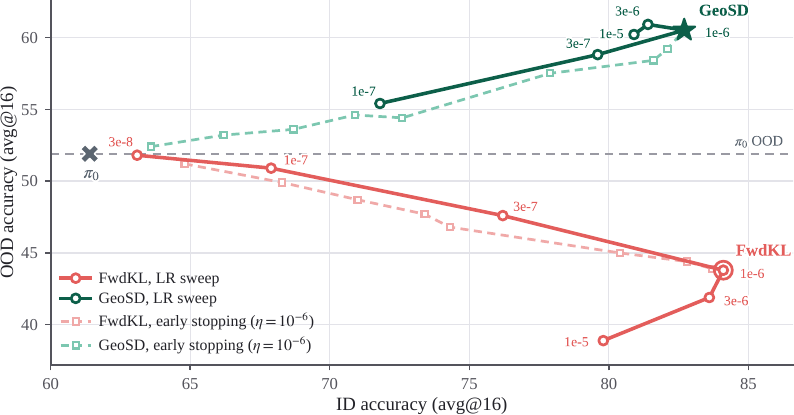}
\caption{\textbf{Learning rate and early stopping.}
ID and OOD accuracy on Qwen3-8B, measured by avg@$16$.
Solid curves connect final checkpoints from the learning-rate
sweep, with rates labeled at each point.
Dashed curves connect intermediate checkpoints at $\eta=10^{-6}$.
The cross marks $\pi_0$, and the horizontal dashed line marks
its OOD accuracy.}
\label{fig:stepsize}
\end{figure}

Figure~\ref{fig:stepsize} shows that reducing the FwdKL learning rate recovers OOD accuracy primarily at the expense of ID gains. At the smallest tested rate, performance approaches that of the base model. Increasing the rate initially improves ID accuracy while lowering OOD accuracy; at the largest tested rate, both are lower than at intermediate rates. The early-stopping trajectory at $\eta=10^{-6}$ shows a similar trade-off and remains below the base model's OOD accuracy.

For \method{}, all plotted learning-rate configurations and intermediate checkpoints improve both ID and OOD accuracy over the base model. Within the tested settings, reducing the learning rate or stopping FwdKL earlier does not recover the combination of ID and OOD gains achieved by \method{}.

\subsection{Attenuation and anchor choice}
\label{app:controls}

We test whether teacher softening or gradient clipping recovers the performance of \method{}, and how fixing the proximal anchor affects ID and OOD accuracy. All controls use Qwen3-8B. The teacher-softening runs follow the FwdKL baseline settings, varying only the teacher temperature over $T\in\{1,1.5,2,3\}$. For the clipping control, we combine forward KL with the Fisher--Rao proximal penalty and K-FAC, clipping each state's matching gradient to the median Hellinger gradient norm measured in Figure~\ref{fig:mechanism}A. We also evaluate Hellinger matching with the Fisher--Rao penalty using clipped AdamW in place of K-FAC. Finally, we compare the default moving anchor with an anchor fixed at $\pi_0$. Other settings follow the corresponding defaults in Table~\ref{tab:hparams}. Table~\ref{tab:controls} reports the results.

Softening the teacher reduces the loss of OOD accuracy but also reduces ID gains. At $T=3$, OOD accuracy reaches $50.6$, still below the base model's $51.9$, while ID accuracy falls by $10.1$ points relative to $T=1$. This resembles the trade-off in the learning-rate sweep, although temperature changes the teacher distribution. Clipped forward KL with the Fisher--Rao penalty and K-FAC exceeds the base model's OOD accuracy but remains $7.3$ points below \method{}. Clipped AdamW achieves $57.3$ OOD accuracy, compared with $56.7$ for plain AdamW and $60.5$ for K-FAC. Fixing the anchor at $\pi_0$ lowers both ID accuracy ($82.7$ to $71.8$) and OOD accuracy ($60.5$ to $55.6$) relative to the moving anchor. Within the tested settings, neither the attenuation controls nor the fixed anchor matches the full method's OOD accuracy.

\begin{table}[t]
\centering
\small
\setlength{\tabcolsep}{5pt}
\caption{Teacher-softening, clipping, and anchor controls on Qwen3-8B.
ID and OOD report avg@16; $\Delta_{\mathrm{OOD}}$ is relative to $\pi_0$ (ID $61.4$, OOD $51.9$).}
\label{tab:controls}
\begin{tabular}{l ccc}
\toprule
Variant & ID & OOD & $\Delta_{\mathrm{OOD}}$ \\
\midrule
FwdKL, $T=1$ & 84.1 & 43.8 & $-8.1$ \\
FwdKL, $T=1.5$ & 81.2 & 46.9 & $-5.0$ \\
FwdKL, $T=2$ & 78.3 & 49.4 & $-2.5$ \\
FwdKL, $T=3$ & 74.0 & 50.6 & $-1.3$ \\
\midrule
FwdKL + per-state clipping, FR + K-FAC
& 80.9 & 53.2 & $+1.3$ \\
\method{} with clipped AdamW
& 82.0 & 57.3 & $+5.4$ \\
\midrule
\method{}, fixed anchor $\pi_0$
& 71.8 & 55.6 & $+3.7$ \\
\method{}, moving anchor
& 82.7 & \textbf{60.5} & $\mathbf{+8.6}$ \\
\bottomrule
\end{tabular}
\end{table}

\subsection{Teacher instantiation}
\label{app:teacher}

The default teacher is the current student conditioned on privileged context (\S\ref{sec:method-setup}). To examine whether slower teacher updates reduce the benefit of proximal regularization, we compare this teacher with a delayed copy refreshed every $125$ updates, an exponential moving average (EMA) of the student parameters with decay $\beta=0.995$, and the frozen base model $\pi_0$. All teachers receive the same form of privileged context. We evaluate each on Qwen3-8B with the full objective and with the proximal penalty removed ($\lambda=0$). The full-objective runs retain the default moving proximal anchor, which is separate from the teacher. Table~\ref{tab:teacher} reports the results.

With the full objective, the three moving teachers achieve OOD accuracies within $0.7$ percentage points of one another. Without the proximal penalty, the delayed and EMA teachers achieve $55.1$ and $56.4$ OOD accuracy, respectively, compared with $53.8$ for the current-student teacher. Nevertheless, removing the proximal penalty lowers OOD accuracy for every teacher choice, including the frozen teacher. Slower teacher updates therefore recover some of the performance lost without proximal regularization, but do not eliminate its benefit in these experiments. Under the full objective, the frozen teacher produces lower ID and OOD accuracy than all three moving teachers.

\begin{table}
\centering
\small
\setlength{\tabcolsep}{5pt}
\caption{Teacher instantiation on Qwen3-8B.
ID and OOD report avg@16.
The teacher and proximal anchor are separate model roles.}
\label{tab:teacher}
\begin{tabular}{l cc cc}
\toprule
& \multicolumn{2}{c}{Full objective}
& \multicolumn{2}{c}{$\lambda=0$} \\
\cmidrule(lr){2-3}
\cmidrule(lr){4-5}
Teacher & ID & OOD & ID & OOD \\
\midrule
Current student & 82.7 & 60.5 & 83.3 & 53.8 \\
Delayed ($K=125$) & 81.9 & 60.8 & 82.4 & 55.1 \\
EMA ($\beta=0.995$) & 82.1 & 61.2 & 82.6 & 56.4 \\
Fixed $\pi_0$ & 76.3 & 57.9 & 77.0 & 55.7 \\
\bottomrule
\end{tabular}
\end{table}

\subsection{Sensitivity to the drift penalty and refresh interval}
\label{app:sensitivity}

We vary the proximal coefficient $\lambda$ and checkpoint-refresh interval $K_{\mathrm{ckpt}}$ separately on Qwen3-8B (Table~\ref{tab:sensitivity}). When varying $\lambda$, we hold $K_{\mathrm{ckpt}}=64$; when varying $K_{\mathrm{ckpt}}$, we hold $\lambda=1$. These sweeps examine sensitivity to the strength of the proximal penalty and the frequency of anchor updates.

OOD accuracy ranges from $59.0$ to $60.5$ for $\lambda\in\{0.3,1,3\}$. Removing the proximal penalty lowers OOD accuracy to $53.8$, while increasing $\lambda$ to $10$ lowers both ID and OOD accuracy relative to the default. For the refresh interval, $K_{\mathrm{ckpt}}\in\{32,64,125\}$ yields OOD accuracies of $59.9$--$60.5$, compared with $58.8$ at $16$ updates, $58.9$ at $250$ updates, and $55.6$ with an anchor fixed at $\pi_0$. Among the tested settings, several penalty strengths and refresh intervals retain most of the default configuration's OOD gain.

\begin{table}
\centering
\small
\setlength{\tabcolsep}{7pt}
\caption{Sensitivity of \method{} to $\lambda$ and $K_{\mathrm{ckpt}}$ on Qwen3-8B (avg@16).
Defaults are marked $^\ast$.
$K_{\mathrm{ckpt}}=\infty$ fixes the anchor at $\pi_0$.}
\label{tab:sensitivity}
\begin{tabular}{l cccccc}
\toprule
$\lambda$ & 0 & 0.1 & 0.3 & $1^\ast$ & 3 & 10 \\
\midrule
OOD & 53.8 & 56.9 & 59.0 & \textbf{60.5} & 60.1 & 58.2 \\
ID  & 83.3 & 83.1 & 82.9 & 82.7 & 81.6 & 78.9 \\
\midrule
$K_{\mathrm{ckpt}}$ & 16 & 32 & $64^\ast$ & 125 & 250 & $\infty$ \\
\midrule
OOD & 58.8 & 59.9 & \textbf{60.5} & 60.2 & 58.9 & 55.6 \\
ID  & 80.2 & 82.0 & 82.7 & 82.9 & 83.0 & 71.8 \\
\bottomrule
\end{tabular}
\end{table}

\section{Compute Analysis}
\label{app:compute}

\begin{table}
\centering\small
\caption{Memory at 8B. The base training footprint
($\approx\!131$\,GB) is shared by all methods. By default, the checkpoint resides on inference workers, leaving $+10.9$\,GB ($\approx\!8\%$) on training workers.
$\times$OPSD is the ratio of total footprint to the OPSD baseline.}
\label{tab:memory_methods}
\begin{tabular}{lccccc}
\toprule
Method & Base train & K-FAC & Checkpoint & Total & $\times$OPSD \\
\midrule
OPSD (standard) & $\approx\!131$\,GB & -- & -- & $\approx\!131$\,GB & 1.00 \\
\method{}       & $\approx\!131$\,GB & $10.9$\,GB & $16.4$\,GB & $\approx\!158$\,GB & 1.21 \\
\bottomrule
\end{tabular}
\end{table}

\begin{table}
\centering\small
\caption{Memory across the Qwen3 scale ladder (GB). \emph{Full EMA} denotes the storage required for unblocked fp32 second-moment EMA factors, shown only for reference and never formed. The implemented K-FAC state uses the default block-diagonal setting per row (fp32, $g{=}16$, up to 8B; bf16, $g{=}32$, at 14B/32B). The last column reports the total method-specific state and its increase over the common training footprint.}
\label{tab:kfac_memory}
\begin{tabular}{l cc c c c c}
\toprule
& & & Reference & \multicolumn{3}{c}{Implemented (default)} \\
\cmidrule(lr){4-4}\cmidrule(lr){5-7}
Model & $g$ & dtype & Full EMA & K-FAC total & $+$ bf16 ckpt & Total (increase) \\
\midrule
Qwen3-1.7B & 16 & fp32 & 17.1  & 2.1  & 3.4  & 5.5\;($+20\%$) \\
Qwen3-4B   & 16 & fp32 & 52.6  & 6.6  & 8.0  & 14.6\;($+23\%$) \\
Qwen3-8B   & 16 & fp32 & 87.3  & 10.9 & 16.4 & 27.3\;($+21\%$) \\
Qwen3-14B  & 32 & bf16 & 183.5 & 5.7  & 29.6 & 35.3\;($+15\%$) \\
Qwen3-32B  & 32 & bf16 & 585.2 & 18.3 & 65.6 & 83.9\;($+16\%$) \\
\bottomrule
\end{tabular}
\end{table}

We express optimizer and scoring costs in \emph{model-pass units}, where one unit is $N B_{\rm tok}$ FLOPs, $N$ is the parameter count, and $B_{\rm tok}$ is the number of student-sequence tokens per update. Under this approximate accounting, a forward pass costs $2$ units and a backward pass $4$. We assign $3$ units to teacher scoring to account for its longer, privileged-context input. \method{} additionally requires a checkpoint forward pass and K-FAC preconditioning. Rollout generation uses the same sampling settings across methods and is included in the runtime measurements but excluded from these unit counts. The Hellinger loss adds negligible computation relative to the model passes. The truncation width $K$ does not change the leading-order model-pass counts because every forward pass computes the full-vocabulary output head before truncation.

\subsection{Memory}

\paragraph{Checkpoint placement.}
The checkpoint is used only for forward passes that produce stop-gradient target logits. It requires no parameter gradients or optimizer state and retains no activations for backward. Its persistent weight storage is $2N$ bytes in bf16 ($16.4$\,GB at 8B), as counted in Table~\ref{tab:memory_methods}; forward-pass buffers are temporary. By default, we place the checkpoint on inference workers, where it scores each rollout after generation. Its retained logits ($<1$\,GB per batch) are transferred to the training workers with the batch, avoiding a second model copy on those workers. The additional persistent state on the training workers is therefore the $10.9$\,GB of K-FAC factors and inverses ($\approx8\%$ of the baseline footprint). We refresh $\theta_{\mathrm{ckpt}}$ every $K_{\mathrm{ckpt}}$ updates using the existing trainer-to-inference weight synchronization path. Each refresh copies the student's current bf16 compute weights without further quantization.

\paragraph{K-FAC memory.}
For each preconditioned linear layer, we store input and output second-moment factors. Summing over the seven projections
$Q,K,V,O,\mathrm{gate},\mathrm{up},\mathrm{down}$ gives
\[
7d^2 + 2d_q^2 + 2d_{kv}^2 + 3d_{\mathrm{ff}}^2,
\qquad
d_q=n_hh,\quad d_{kv}=n_{kv}h,
\]
entries per transformer block before applying the block-diagonal approximation. Here, $n_h$ and $n_{kv}$ are the numbers of query and key--value heads, respectively, and $h$ is the head dimension. The expression reduces to $9d^2+2d_{kv}^2+3d_{\mathrm{ff}}^2$ when $n_hh=d$, which holds for all listed models except Qwen3-4B and Qwen3-32B. Partitioning each factor into $g$ equally sized diagonal blocks reduces factor storage by a factor of $g$ and the arithmetic cost of dense inversion by a factor of $g^2$. The reported K-FAC storage includes both the EMA factors and their damped inverses, stored at the precision specified in Table~\ref{tab:kfac_memory}, and is therefore twice the blocked factor storage. These totals exclude temporary workspaces.

Across the Qwen3 sizes, persistent K-FAC storage ranges from $2.1$\,GB at 1.7B to $18.3$\,GB at 32B. For reference, full unblocked fp32 factors would require up to $585$\,GB at 32B; these are never formed. The bf16 checkpoint adds $2N$ bytes, ranging from $3.4$\,GB at 1.7B to $65.6$\,GB at 32B. At 8B, K-FAC state and checkpoint weights together add $27.3$\,GB in total, or $21\%$ of the $\approx\!131$\,GB baseline training footprint (Table~\ref{tab:memory_methods}). With the checkpoint on inference workers, only the $10.9$\,GB of persistent K-FAC state resides on the training workers, an increase of approximately $8\%$ over that baseline.

\subsection{Runtime}

On H200 GPUs, an 8B baseline run requires approximately $7$ GPU-h for training and $28$ GPU-h for evaluation (avg@$16$/pass@$16$). The checkpoint forward pass and K-FAC operations increase \method{}'s per-update optimizer and scoring FLOPs by $26\%$ ($1.26\times$) relative to standard OPSD. The measured wall-clock increase is approximately $10\%$ ($1.10\times$), with checkpoint scoring on inference workers overlapping rollout generation. Generation dominates runtime and uses the same decoding settings across methods. Across the nine post-training methods, training and evaluation require approximately $63$ and $252$ GPU-h, respectively. Including approximately $20$ GPU-h for offline construction of the shared privileged-solution pool, the comparison costs approximately $3170$ GPU-h for $10$ runs with different seeds.
\section{Limitations}
\label{app:limitations}

\paragraph{Learning beyond the base model.}
\method{} encourages small, gradual changes to the student's predictions. This may hurt learning when a correct continuation initially has very low student probability. Overlap weighting attenuates the teacher's preference and the proximal penalty discourages large departures from recent predictions. Although this can protect the student from harmful changes in settings with privileged context, both terms can also hold back useful corrections.
We also exclude training problems for which the base model produces no verified solution within the sampling budget. It remains open whether \method{} can learn from solutions to problems the base model cannot solve within that budget.

\paragraph{Limits of geometric control.}
The moving checkpoint discourages local predictive drift but does not impose a hard bound on drift from the initial model. The checkpoint and student can drift together over longer training runs. The penalty is also evaluated only on sampled training prefixes, leaving changes at other states unconstrained by it.

\paragraph{Scope and interpretation of the evidence.}
We evaluate mathematics across model families and scales, but code generation and teacher ablations are limited to Qwen3-8B. The concentration and false-consensus analyses cover mathematics only. They associate these behaviors with OOD performance but do not establish that preserving alternative token mass causes the accuracy gains.
We have not tested whether the same explanation applies to code or whether \method{} preserves performance on tasks beyond the evaluated benchmarks.
\section{AI use statement}

We used an LLM to edit text at the sentence level to improve clarity, concision, and grammar. We also used it to find related work and explore the literature, check existing mathematical derivations, and assist with plotting code and scripts for running experiments.
\section{Reproducibility statement}
Appendix~\ref{app:exp-details} documents the model families and checkpoint variants, datasets, data splits, and construction of the verified privileged-context pool, together with training and evaluation protocols, baseline implementations, hyperparameter selection, and hardware. Model architectures and hyperparameters are summarized in Tables~\ref{tab:archs} and~\ref{tab:hparams}, and the K-FAC implementation is detailed in Appendix~\ref{app:exp-kfac}. Appendix~\ref{app:compute} reports computational and memory costs (Tables~\ref{tab:memory_methods} and~\ref{tab:kfac_memory}).
Our code is available at \url{https://github.com/josipjukic/geosd/}.

\end{document}